\documentclass[10pt,conference]{IEEEtran}

\usepackage{amsmath,amssymb,amsfonts,mathtools,bm}
\usepackage{amsthm}
\usepackage{stmaryrd}
\usepackage{booktabs,tabularx,array,makecell,multirow}
\usepackage{graphicx}
\usepackage{algorithm}
\usepackage{algpseudocode}
\usepackage[normalem]{ulem}
\usepackage{enumitem}
\usepackage{xcolor}
\usepackage{adjustbox}
\usepackage{placeins}
\usepackage{flushend}
\usepackage{microtype}
\usepackage[english]{babel}
\makeatletter
\let\l@en\l@english
\makeatother
\usepackage[numbers,sort&compress]{natbib}
\usepackage[colorlinks=true,linkcolor=blue,citecolor=blue,urlcolor=blue]{hyperref}

\makeatletter
\apptocmd{\thebibliography}{\small}{}{}
\makeatother

\newenvironment{ieeewidetable}[1]
  {\def\ieeetablewidth{\textwidth}}
  {}

\graphicspath{{assets/}{figures/}}

\newcommand{\R}{\mathbb{R}}
\newcommand{\N}{\mathbb{N}}

\newcommand{\U}{\mathcal{U}}

\newcommand{\Obs}{\mathcal{O}}

\DeclareMathOperator{\dist}{dist}

\newcommand{\norm}[1]{\left\lVert#1\right\rVert}
\newcommand{\set}[1]{\{#1\}}

\newcommand{\CSymPlan}{CSymPlan}
\newcommand{\ee}{end-effector}

\theoremstyle{plain}

\newtheorem{proposition}{Proposition}
\newtheorem{problem}{Problem}

\theoremstyle{definition}
\newtheorem{assumption}{Assumption}

\theoremstyle{remark}
\newtheorem{remark}{Remark}

\title{CSymPlan: Certified Symbolic Planning and Control for High-DOF Manipulators}

\author{%
\IEEEauthorblockN{%
Aditya Narendra\IEEEauthorrefmark{1},
Ashok Kumar Saini\IEEEauthorrefmark{1},
Mahathi Anand\IEEEauthorrefmark{2},
Mahmoud Khaled\IEEEauthorrefmark{3},\\
Fares J. Abu-Dakka\IEEEauthorrefmark{4}, and
Abdalla Swikir\IEEEauthorrefmark{1}}
\IEEEauthorblockA{%
\IEEEauthorrefmark{1}Mohamed bin Zayed University of Artificial Intelligence (MBZUAI), Abu Dhabi, United Arab Emirates\\
\IEEEauthorrefmark{2}Learning Systems and Robotics Lab, Technical University of Munich, Germany\\
\IEEEauthorrefmark{3}Ludwig Maximilian University of Munich, Munich, Germany\\
\IEEEauthorrefmark{4}New York University Abu Dhabi, Abu Dhabi, United Arab Emirates\\
Corresponding author: Aditya Narendra (aditya.narendra@mbzuai.ac.ae)}
}

\hypersetup{
  pdftitle={CSymPlan: Certified Symbolic Planning and Control for High-DOF Manipulators},
  pdfauthor={Aditya Narendra, Ashok Kumar Saini, Mahathi Anand, Mahmoud Khaled, Fares J. Abu-Dakka, Abdalla Swikir}
}

\begin{document}
\maketitle
\pagestyle{plain}
\thispagestyle{plain}

\begin{abstract}
Robot manipulators are commonly engineered around a decoupled motion-generation stack: a planner computes a collision-free path and a lower-level controller tracks the resulting reference. This separation is computationally convenient, but it can produce references that are difficult to execute under actuator limits, tracking error, model mismatch, and small obstacle clearances. We present \CSymPlan{}, a certified symbolic planning and control framework for high-DOF manipulators with two complementary implementations: an offline implementation that precomputes certified reach-avoid feedback policies for known workspaces; and an online implementation that synthesizes or updates symbolic policies at runtime from changing task and perception information using parallelization. The offline implementation reduces the manipulator dynamics to a sampled perturbed double-integrator model in operational space through feedback linearization, treats torque-realization errors, modeling inaccuracies, and measurement uncertainty as bounded disturbances, and refines the synthesized symbolic policy to the Franka FR3 through a quantization--lookup--torque realization pipeline. The online implementation uses the same abstraction and refinement interface, but replaces the precomputed policy table with a runtime pFaces request--synthesis--execution loop. In randomized simulated benchmarks and perception-driven Franka FR3 experiments, both implementations complete reach-avoid tasks with zero safety violations; whenever no certified action exists, the robot holds, replans, or stops safely instead of executing an uncertified command.
\end{abstract}


\section{Introduction}
\label{sec:introduction}

\begin{figure*}[t]
    \centering
    \includegraphics[width=1\textwidth]{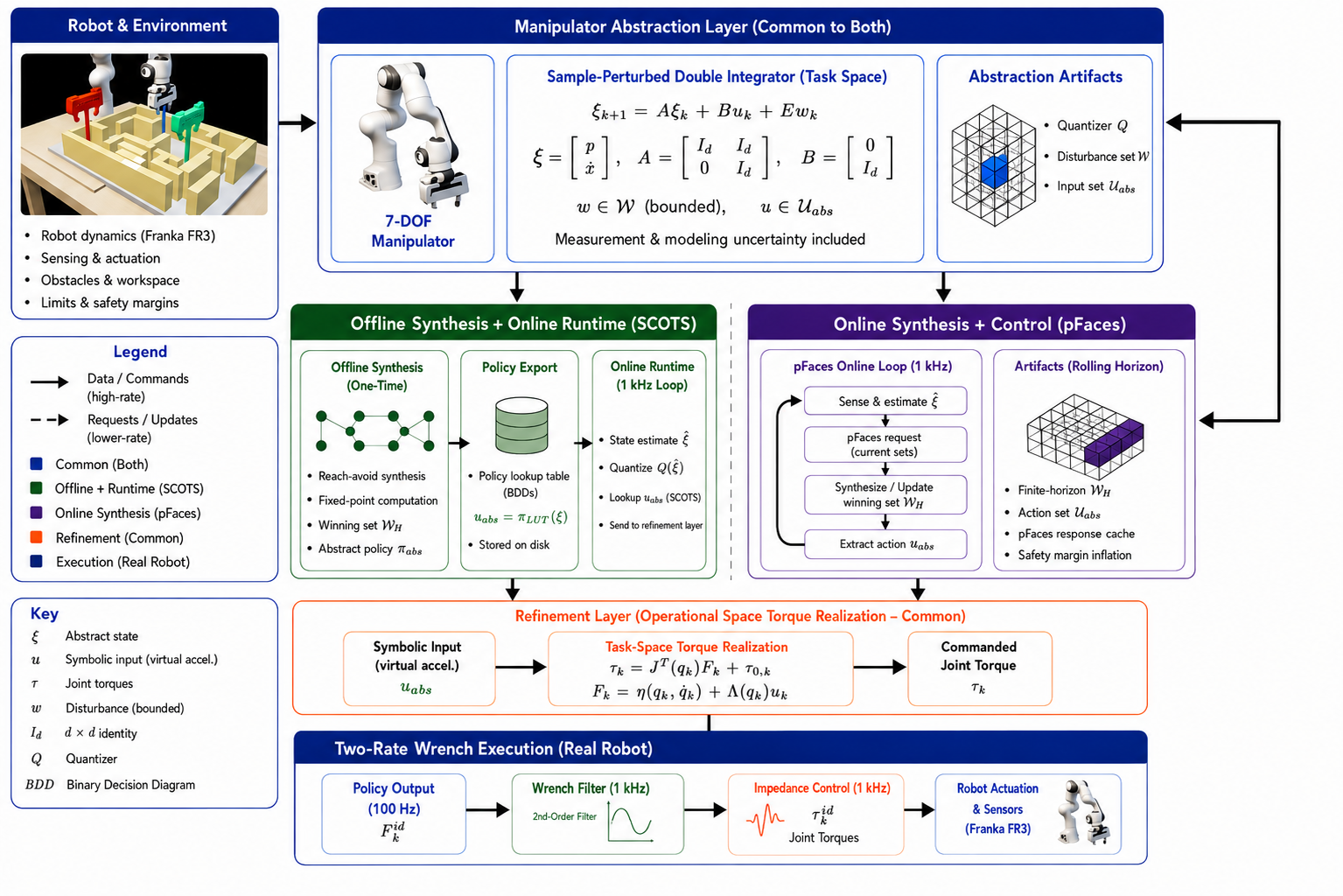}
    \caption{Overview of \CSymPlan{}. The offline implementation synthesizes a certified reach-avoid policy from a finite abstraction of sampled perturbed task-space dynamics and executes it through a quantization--lookup--torque pipeline. The online implementation uses the same abstraction and torque-realization interface, but calls pFaces online mode at runtime to synthesize or refresh the symbolic policy for the current target and obstacle sets.}
    \label{fig:csymplan_overview}
\end{figure*}

Precise motion generation for robot manipulators~\citep{curobo} is central to applications ranging from industrial pick-and-place to human--robot interaction and other safety-critical manipulation settings \citep{khatib_robot_1997}. 
Conventional architectures separate motion generation into two layers. A high-level planner first generates a geometric path or reference trajectory \citep{lavalle_planning_2006}; a low-level feedback controller then attempts to track that reference on the physical robot \citep{ajwad_systematic_2015}. Although this modular design has been successful in practice, it also creates a gap between planning and execution. A path that is collision-free in the planner may become unsafe after tracking errors, actuator saturation, unmodeled dynamics, or sensing errors are introduced at runtime.

This gap is most visible in cluttered environments, where the clearance around obstacles can be comparable to the tracking error of the manipulator. In such settings, a geometric planner may return a feasible-looking path that is dynamically difficult to track, while a controller may stabilize the robot locally without guaranteeing that all future closed-loop states remain inside a safe set. Recent integrated planning-and-control approaches, including kinodynamic planning, model predictive control, model predictive path integral control, and reinforcement-learning-based controllers, reduce this mismatch by considering dynamics during decision making \citep{lavalle2001randomized,mayne_constrained_2000,williams_mppi_2017,kober2013reinforcement,schulman2017ppo,haarnoja2018sac}. However, these methods often rely on local optimization, stochastic sampling, learned approximations, or computationally intensive online operations; most do not provide a direct closed-loop reach-avoid certificate for the executed manipulator trajectory \citep{tabuada2009verification,pola2008symbolic,zamani_symbolic_2012,rungger_scots_2016}.

To close this gap, we propose \CSymPlan{}, which is, to the best of our knowledge, the first unified symbolic planning-and-control framework for high-DOF manipulators demonstrated on torque-controlled hardware. Here, the word symbolic is used in the control-theoretic sense: symbols denote finitely many discretized cells of continuous state and input spaces. Consequently, rather than synthesizing a purely geometric path, \CSymPlan{} constructs a finite-state abstraction of the task-space closed-loop dynamics and synthesizes a feedback policy that satisfies reachability and safety specifications by construction. The synthesized policy maps each abstract/symbolic state to one or more admissible control inputs whose successors remain inside the certified winning set.

The main technical challenge is scalability. A direct abstraction of a 7-DOF manipulator in joint position and velocity space is not tractable with uniform discretization. \CSymPlan{} avoids this bottleneck by abstracting the translational \ee{} dynamics in operational space. Through task-space feedback linearization, the manipulator is represented as a sampled perturbed double integrator in position and velocity. The abstraction explicitly incorporates bounded disturbances and measurement uncertainty. The resulting symbolic action is interpreted as a virtual Cartesian acceleration and refined to joint torques through an operational-space torque controller with null-space damping and posture regulation. 

A policy synthesized offline is, however, bound to the workspace, obstacle set, and goal for which it was computed: any change invalidates the certificate, so offline synthesis suits static, known workspaces but not tasks whose goals arrive at runtime or whose obstacles are perceived during execution. Moving synthesis into the control loop raises three challenges: fixed-point synthesis is orders of magnitude slower than the $1\,\mathrm{kHz}$ servo rate; a certificate holds only for the state, safe set, and horizon for which it was computed, so certified commands become stale as the robot and environment evolve; and rebuilding the abstraction at every replanning instant is intractable. How the online implementation meets these challenges is summarized in the contributions below and detailed in Section~\ref{sec:online_implementation}. Figure~\ref{fig:csymplan_overview} summarizes the full offline--online pipeline.

The contributions of this paper are organized around two implementations of the same certified planning-and-control principle.
\begin{enumerate}[leftmargin=*]
\item \textbf{Task-space abstraction for high-DOF arms.} We reduce a high-dimensional manipulator to a tractable Cartesian double-integrator abstraction using operational-space feedback linearization, while retaining a torque-level refinement layer for execution on a Franka FR3. 

\item \textbf{Offline certified symbolic synthesis for manipulation.} We formulate manipulation reach-avoid tasks as symbolic feedback synthesis problems over sampled perturbed task-space dynamics, yielding correct-by-construction policies under explicit disturbance and measurement bounds.

\item \textbf{Real-time execution of precomputed policies.}  We implement a quantization--lookup--torque pipeline that executes the offline-synthesized policy at the robot servo rate and demonstrate hardware operation on reach-avoid manipulation tasks. 

\item \textbf{Online symbolic planning and control.} We extend the framework to runtime-changing tasks and perception-driven obstacle sets: the grids and transition relation are fixed once, finite-horizon reach-avoid winning sets are recomputed on demand by parallelized synthesis (pFaces~\citep{pfaces}), and the robot executes buffered certified acceleration commands, with stale-command rejection, through the same torque-realization layer.

\item \textbf{Simulation and hardware evaluation protocol.} We define PyBullet and Isaac Sim~\citep{isaac_sim} reach-avoid benchmarks and real Franka FR3 experiments used to evaluate success rate, safety violations, safe resolution, time-to-goal, path length, smoothness, effort, and online synthesis time against planning, receding-horizon, safety-filter, and learned baselines.

Figure~\ref{fig:csymplan_overview} summarizes the full offline--online \CSymPlan{} pipeline.

\end{enumerate}

\section{Related Work}
\label{sec:related_work}

\subsection{Geometric and sampling-based planning}

Classical graph-search planners such as A$^\star$ and Dijkstra's algorithm operate on discrete representations of the environment and provide useful completeness properties on the chosen graph \citep{candra_dijkstras_2020}. Sampling-based planners, including rapidly-exploring random trees and probabilistic roadmaps, extend motion planning to high-dimensional spaces by searching continuous configuration spaces through samples \citep{RRT_star,orthey_sampling-based_2024}. These methods are practical and often effective for kinematic planning, but they typically return paths rather than closed-loop controllers. If the robot cannot track the planned path with sufficient accuracy, the execution may violate safety constraints despite the geometric validity of the reference.

\subsection{Kinodynamic and optimization-based planning}

Kinodynamic and optimization-based planners reduce the gap between planning and execution by incorporating system dynamics, actuator constraints, or trajectory regularity directly into motion generation \citep{lavalle2001randomized,opt_planning}. Receding-horizon methods such as model predictive control (MPC) repeatedly optimize a finite-horizon control problem from the current state and commonly use a reference trajectory or target state in the objective, thereby combining trajectory tracking with online constraint-aware control \citep{mpc_planning,mpc_planning2}. Model predictive path integral (MPPI) control similarly computes receding-horizon actions online, but uses stochastic trajectory sampling and cost-weighted updates rather than deterministic numerical optimization \citep{williams_mppi_2017}. These feedback formulations can respond to disturbances and model deviations more naturally than a strictly decoupled plan--then--track pipeline.

Reinforcement-learning-based controllers and neural approximations of MPC can further reduce online computation or improve performance for task distributions encountered during training \citep{rl_planning}. However, the guarantees provided by these methods depend strongly on their particular formulation. Nominal finite-horizon MPC, for example, does not by itself imply recursive feasibility or closed-loop constraint satisfaction under bounded disturbances; such guarantees generally require additional ingredients such as robust invariant sets, terminal constraints or costs, constraint tightening, or robust/tube MPC constructions \citep{mayne_constrained_2000,rawlings_mpc_2017}. Likewise, the sampling procedure in MPPI provides an approximate solution to a stochastic finite-horizon optimal-control problem rather than an exhaustive verification of all possible successors from a region of the state space \citep{williams_mppi_2017}. Learned controllers inherit no formal reach-avoid guarantee solely from training performance unless an additional verification or certified safety mechanism is introduced. Consequently, while these approaches can provide excellent empirical closed-loop performance, their standard formulations do not directly produce the type of set-wise reach-avoid certificate considered here: a feedback policy for which every admissible successor, for every state represented by a certified abstract cell and every disturbance within a prescribed bound, is guaranteed to remain safe and eventually reach the goal.

\subsection{Symbolic control}


Symbolic control constructs finite transition systems that over-approximate continuous or hybrid dynamics and synthesizes controllers for temporal, reachability, or safety specifications \citep{tabuada2009verification,belta_symbolic_2007}. Feedback refinement relations connect the abstract and concrete systems so that, under a sound abstraction, the refined controller inherits the abstract guarantees \citep{symbolic_feedback,rungger_scots_2016}.

Although the term ``symbolic'' is also used in task and motion planning (TAMP), it has a different meaning. TAMP typically uses symbols to represent semantic predicates, objects, and high-level actions, while continuous planners determine configurations or trajectories that realize those actions \citep{wolfe_combined_2010,kaelbling_hierarchical_2011,srivastava_combined_2014}. In contrast, symbolic control uses symbols to represent regions of the continuous state and input spaces and synthesizes a feedback policy over these regions. Thus, TAMP primarily addresses task-level sequencing and continuous feasibility, whereas symbolic control focuses on certifying closed-loop dynamical behavior. The two approaches are complementary: a TAMP layer could provide high-level manipulation goals that are executed by certified symbolic controllers.

Symbolic-control methods have been applied to systems such as mobile robots and autonomous vehicles \citep{fainekos_temporal_2009,symbolic_vehicle}, but their main limitation is state-space explosion. \CSymPlan{} addresses this challenge for high-DOF manipulators by synthesizing over a lower-dimensional operational-space model rather than the full joint-space state.

\begin{figure*}[t]
    \centering
    \includegraphics[width=0.98\textwidth]{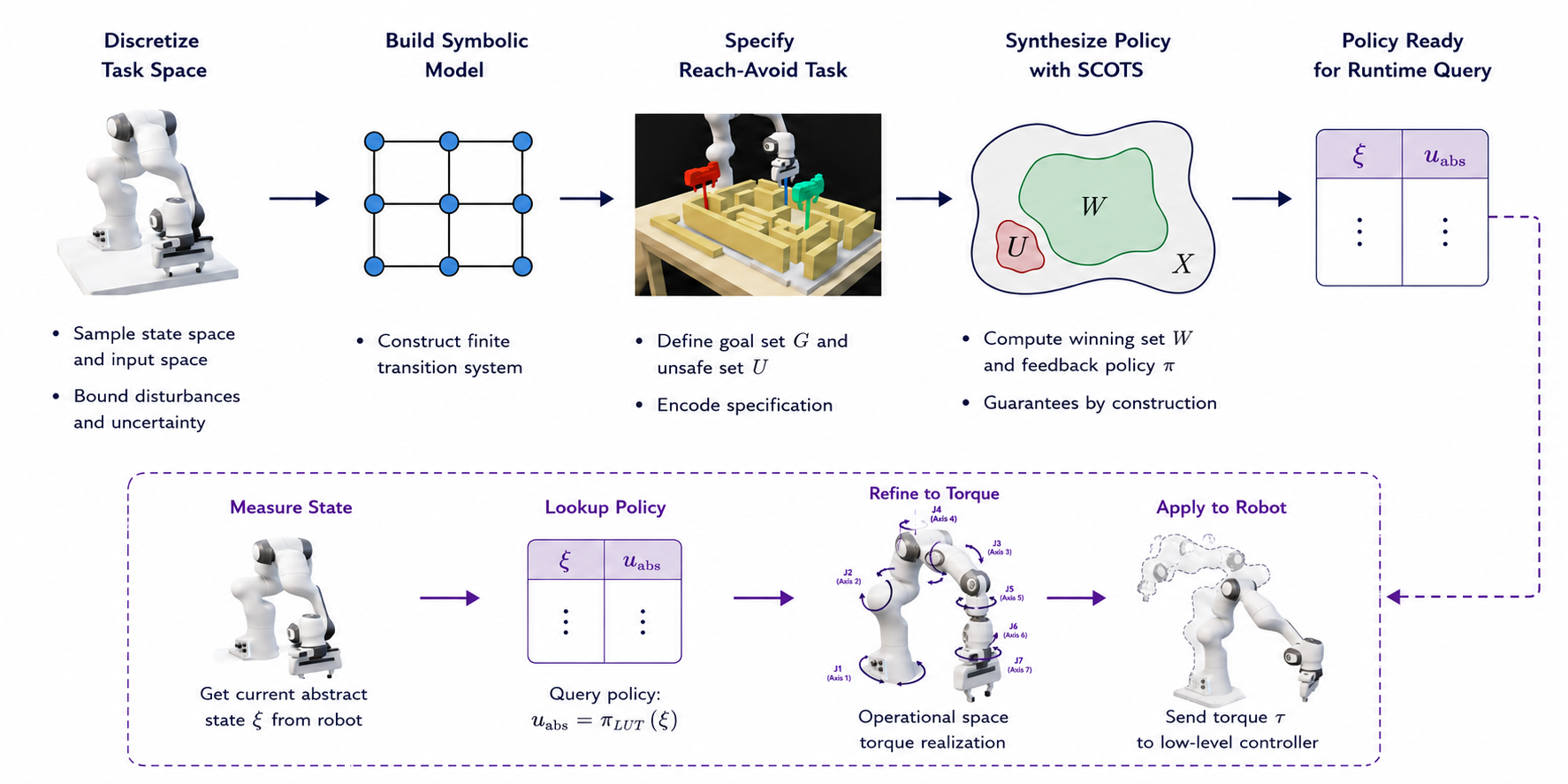}
    \caption{Offline \CSymPlan{} architecture. The upper row shows the one-time synthesis pipeline: task-space discretization, finite symbolic model construction, reach-avoid task specification, SCOTS-based policy synthesis, and export of a policy table for runtime query. The lower row shows the real-time execution pipeline: the robot state is measured and quantized, the precomputed policy is queried, the selected abstract input is refined through operational-space torque realization, and the resulting torque command is applied to the robot.}
    \label{fig:offline_architecture}
\end{figure*}

\section{Background and Problem Formulation}
\label{sec:background}

\subsection{Notation}

Let $\R$, $\R_{\ge 0}$, and $\N$ denote the sets of real numbers, non-negative real numbers, and natural numbers, respectively. For $x,\hat x\in\R^n$, $\dist(x,\hat x)=\norm{x-\hat x}$ denotes Euclidean distance. For $x\in\R^n$ and $A\subset\R^n$, $\dist(x,A)=\inf_{a\in A}\norm{x-a}$. A set-valued map from $A$ to $B$ is denoted $f:A\rightrightarrows B$. Infinite sequences are written in bold, for example $\bm{x}=(x_0,x_1,\ldots)$, and finite truncations are denoted by a subscript such as $\bm{x}_T$.

\subsection{Manipulator dynamics in operational space}
\label{sec:op_space_dynamics}

Consider an $n_q$-DOF manipulator with joint configuration $q\in\R^{n_q}$. Its rigid-body dynamics are
\begin{equation}
    M(q)\ddot q+C(q,\dot q)\dot q+g(q)=\tau+\tau_{\mathrm{ext}},
    \label{eq:manip_joint_dyn}
\end{equation}
where $M(q)$ is the inertia matrix, $C(q,\dot q)\dot q$ contains Coriolis and centrifugal terms, $g(q)$ is gravity, $\tau$ is the commanded joint torque, and $\tau_{\mathrm{ext}}$ represents external torques. Let the translational \ee{} position be $x=h(q)\in\R^d$, with $d\in\{2,3\}$, and let $J(q)=\partial h/\partial q$ be the corresponding Jacobian. Then
\begin{equation*}
    \dot x=J(q)\dot q,\qquad \ddot x=J(q)\ddot q+\dot J(q,\dot q)\dot q.
\end{equation*}
The operational-space dynamics can be written as \citep{khatib_os}
\begin{equation}
    \Lambda(q)\ddot x+\eta(q,\dot q)=F+F_{\mathrm{ext}},
    \label{eq:opspace_dyn}
\end{equation}
where $\Lambda(q)=(J M^{-1}J^\top)^{-1}$ is the operational-space inertia, $\eta(q,\dot q)$ groups task-space bias terms, $F\in\R^d$ is the commanded task-space wrench for the translational task, and $F_{\mathrm{ext}}$ collects external task-space effects.

\subsection{Sampled perturbed control systems}
\label{sec:sampled_perturbed}

A sampled perturbed control system is described by a tuple $S=(X,U,\mathcal{F})$, where $X\subseteq\R^n$ is the state set, $U\subseteq\R^m$ is the input set, and $\mathcal{F}:X\times U\rightrightarrows X$ is a set-valued transition map. For sampling period $s>0$, $\mathcal{F}$ is induced by the sampled behavior of the perturbed continuous-time differential inclusion
\begin{equation}
    \dot\xi\in f(\xi,u)\oplus\llbracket-w,w\rrbracket,
    \label{eq:symbolic_dyn}
\end{equation}
where $\xi\in X$, $u\in U$, and $w\in\R^n_{\ge0}$ bounds the component-wise disturbance set
\begin{equation*}
    \llbracket-w,w\rrbracket=[-w_1,w_1]\times\cdots\times[-w_n,w_n].
\end{equation*}
Measurement uncertainty is represented by \eqref{eq:symbolic_meas}:
\begin{equation}
    P(x)=x\oplus\llbracket-z,z\rrbracket,
    \label{eq:symbolic_meas}
\end{equation}
with $z\in\R^n_{\ge0}$. A behavior of $S$ is an infinite sequence $(\bm{x},\bm{u})=(x_0,u_0,x_1,u_1,\ldots)$ such that $x_{k+1}\in \mathcal{F}(x_k,u_k)$ for all $k\in\N$.

\subsection{Reachability, invariance, and reach-avoid specifications}
\label{sec:specifications}

A specification is a desired subset of possible state sequences. In this work we consider invariance, reachability, and reach-avoid specifications. Given a safe set $X_{\mathrm{safe}}$, invariance requires the system to remain safe for all time:
\begin{equation}
    \Sigma_{\mathrm{inv}}=\set{\bm{x}\mid x_0\in X_{\mathrm{safe}}\Rightarrow x_k\in X_{\mathrm{safe}},\ \forall k\in\N}.
    \label{eq:inv}
\end{equation}
Given an initial set $X_0$, goal set $X_{\mathrm{goal}}$, and horizon $T$, reachability requires the system to reach the goal within the horizon:
\begin{equation}
    \Sigma_{\mathrm{reach}}=\set{\bm{x}\mid x_0\in X_0\Rightarrow \exists k\in\{0,\ldots,T\}:x_k\in X_{\mathrm{goal}}}.
    \label{eq:reach}
\end{equation}
The reach-avoid specification in \eqref{eq:reach_avoid} combines the invariance and reachability specifications in \eqref{eq:inv} and \eqref{eq:reach}:
\begin{align}
    \Sigma_{\mathrm{RA}}=\set{\bm{x}\mid x_0\in X_0\Rightarrow
    &\exists k\in\{0,\ldots,T\}:x_k\in X_{\mathrm{goal}}, \notag\\
    &x_i\in X_{\mathrm{safe}},\ \forall i\le k}.
    \label{eq:reach_avoid}
\end{align}
A controller satisfies a specification if the closed-loop behaviors of the system under that controller are contained in the specification set.

\subsection{Symbolic controller synthesis}
\label{sec:symbolic_synthesis}

Symbolic synthesis proceeds by constructing a finite abstraction $S_{\mathrm{abs}}=(X_{\mathrm{abs}},U_{\mathrm{abs}},\mathcal{F}_{\mathrm{abs}})$ that over-approximates the sampled perturbed system $S$. In the implementation, the abstraction is connected to the concrete system by a quantizer $Q:X\to X_{\mathrm{abs}}$. Equivalently, the graph of the quantizer,
\begin{equation*}
    R_Q=\set{(x,\hat x)\in X\times X_{\mathrm{abs}}\mid \hat x=Q(x)},
\end{equation*}
can be viewed as the feedback-refinement relation. Assuming $U_{\mathrm{abs}}\subseteq U$, refinement requires that every admissible symbolic input is admissible for the concrete plant and that all quantized concrete successors are included among the abstract successors. In particular, for every $(x,\hat x)\in R_Q$,
\begin{enumerate}[leftmargin=*]
\item $U_{\mathrm{abs}}(\hat x)\subseteq U(x)$; and
\item for all $u\in U_{\mathrm{abs}}(\hat x)$, $Q(\mathcal{F}(x,u))\subseteq \mathcal{F}_{\mathrm{abs}}(\hat x,u)$, where $Q(\mathcal{F}(x,u))=\set{Q(x^+)\mid x^+\in \mathcal{F}(x,u)}$.
\end{enumerate}
The abstract controller is then computed by fixed-point algorithms for reachability and invariance \citep{rungger_scots_2016}. If the abstraction is sound and the safe and goal sets are approximated conservatively, the refined controller inherits the abstract guarantee.

\section{Methodology}
\label{sec:methodology}

This section presents the common mathematical model used by both the offline and online versions of \CSymPlan{}. The offline implementation precomputes a fixed symbolic controller for a given workspace, obstacle set, and goal. The online implementation retains the same symbolic state and input grids and transition relation, but recomputes the finite-horizon reach-avoid winning set and corresponding admissible control actions at runtime from the current measured state, target set, and obstacle set. In our implementation, these online fixed-point computations are executed in parallel using pFaces~\citep{pfaces}.

\subsection{Manipulator reduction to a sampled perturbed double integrator}
\label{sec:robot_sampled}

\CSymPlan{} begins with the operational-space model in \eqref{eq:opspace_dyn}. We choose the task-space command
\begin{equation}
    F=\eta(q,\dot q)+\Lambda(q)u,
    \label{eq:feedback_lin_force}
\end{equation}
where $u\in\R^d$ is a virtual input interpreted as desired \ee{} acceleration. Substituting \eqref{eq:feedback_lin_force} into \eqref{eq:opspace_dyn} gives
\begin{equation}
    \ddot x=u+\Lambda(q)^{-1}F_{\mathrm{ext}}.
    \label{eq:ddx_virtual}
\end{equation}
The residual term in \eqref{eq:ddx_virtual}, $\Lambda(q)^{-1}F_{\mathrm{ext}}$, together with modeling error, discretization error, measurement error, and torque-realization error, is modeled as a bounded additive disturbance. Defining the task-space state $\xi=[x^\top,v^\top]^\top$, with $v=\dot x$, yields the perturbed double-integrator model
\begin{equation}
    \dot\xi=
    \begin{bmatrix}
0&I_d\\ 0&0
    \end{bmatrix}\xi+
    \begin{bmatrix}
0\\ I_d
    \end{bmatrix}u
\oplus\llbracket-w,w\rrbracket.
    \label{eq:sp_mani_sys}
\end{equation}
This is an instance of the perturbed differential inclusion \eqref{eq:symbolic_dyn} with linear drift. The input set $u\in\U\subset\R^d$ is compact and is selected to respect actuation limits after torque realization. The system is sampled with period $s$ under zero-order-hold inputs. In practice, the disturbance bound $w$ and measurement bound $z$ are calibrated conservatively from residual-acceleration and state-estimation error data recorded while running the torque-realization layer of Section~\ref{sec:implementation} on representative motions.

\subsection{Safe and goal sets}
\label{sec:safe_goal_sets}

Let $X_{\mathrm{ws}}\subset\R^d$ denote the admissible Cartesian workspace and let $\Obs\subset X_{\mathrm{ws}}$ be the union of obstacle regions. For a safety distance $d_{\mathrm{safe}}>0$, define the inflated obstacle set
\begin{equation}
    \Obs^{d_{\mathrm{safe}}}=\set{x\in\R^d\mid \dist(x,\Obs)\le d_{\mathrm{safe}}}.
    \label{eq:obs}
\end{equation}
Using the inflated obstacle set in \eqref{eq:obs}, the position-level safe set is
\begin{equation*}
    X^x_{\mathrm{safe}}=X_{\mathrm{ws}}\setminus\Obs^{d_{\mathrm{safe}}}.
\end{equation*}
Lifting this set to the full task-space state gives
\begin{equation}
    X_{\mathrm{safe}}=\set{\xi=[x^\top,v^\top]^\top\in\R^{2d}\mid x\in X^x_{\mathrm{safe}},\ v\in V},
    \label{eq:safe_set}
\end{equation}
where $V\subset\R^d$ is a compact velocity set. The safe set is stated over the translational \ee{} position; accordingly, the margin $d_{\mathrm{safe}}$ is chosen to cover the \ee{} and tool geometry in addition to the tracking and measurement uncertainty already captured by $w$ and $z$. Certification of whole-arm collision avoidance is outside the scope of the present abstraction; extensions beyond translational task-space certification are discussed in Section~\ref{sec:future_joint_space}. For a target position $x_T\in X^x_{\mathrm{safe}}$ and position tolerance $d_{\mathrm{tol}}$, define
\begin{equation*}
    X^x_{\mathrm{goal}}=\set{x\in X_{\mathrm{ws}}\mid \dist(x,x_T)\le d_{\mathrm{tol}}}.
\end{equation*}
Because the robot should come to rest at the goal, the full target set is
\begin{equation}
    X_{\mathrm{goal}}=\set{\xi=[x^\top,v^\top]^\top\in\R^{2d}\mid x\in X^x_{\mathrm{goal}},\ \norm{v}_{\infty}\le v_{\mathrm{tol}}}.
    \label{eq:goal_set}
\end{equation}
Sequential waypoint tasks are handled by synthesizing a separate reach-avoid policy for each waypoint and switching to the next policy once the current goal set has been reached and held for a specified number of control ticks.

\begin{problem}[Certified reach-avoid manipulation]\label{prob:reach_avoid}
Given the manipulator dynamics \eqref{eq:manip_joint_dyn} with translational \ee{} output $x=h(q)$, workspace $X_{\mathrm{ws}}$, obstacle set $\Obs$, disturbance and measurement bounds $w$ and $z$, an initial set $X_0\subseteq X_{\mathrm{safe}}$, and the safe and goal sets \eqref{eq:safe_set} and \eqref{eq:goal_set}, synthesize a joint-torque feedback controller such that every closed-loop behavior of the sampled perturbed task-space system \eqref{eq:sp_mani_sys} satisfies the reach-avoid specification \eqref{eq:reach_avoid}.
\end{problem}
The offline implementation of Section~\ref{sec:offline_implementation} solves Problem~\ref{prob:reach_avoid} once per scene by precomputing a policy over the full abstraction; the online implementation of Section~\ref{sec:online_implementation} solves a finite-horizon instance of Problem~\ref{prob:reach_avoid} at each replanning step for the currently perceived sets.

\section{Implementation}
\label{sec:implementation}

\subsection{Offline implementation}
\label{sec:offline_implementation}
Figure~\ref{fig:offline_architecture} shows the offline architecture: a one-time synthesis pipeline that discretizes the task space, builds the finite symbolic model, and exports a reach-avoid policy table, followed by a real-time execution pipeline that queries the table at the robot servo rate.

\subsubsection{Offline symbolic synthesis}
\label{sec:offline_synthesis}

For a fixed workspace, obstacle set, and goal set, \CSymPlan{} first constructs a finite abstraction of \eqref{eq:sp_mani_sys}. The continuous set $X=X_{\mathrm{ws}}\times V$ is discretized into a uniform grid $X_{\mathrm{abs}}$, and the admissible acceleration set $\U$ is discretized into a finite input set $U_{\mathrm{abs}}$. For every abstract state--input pair, the transition relation $\mathcal{F}_{\mathrm{abs}}$ is computed by over-approximating all reachable sampled successors under the disturbance and measurement bounds. The abstract safe set is selected conservatively so that each represented concrete cell lies in $X_{\mathrm{safe}}$, and the abstract goal set is selected conservatively so that each represented concrete cell lies in $X_{\mathrm{goal}}$.

A fixed-point computation then yields a winning set and a symbolic controller. The resulting controller is exported as a binary decision diagram or lookup table. At runtime, policy evaluation is therefore constant time: the measured state is quantized, the symbolic action is looked up, and the corresponding Cartesian acceleration is refined to torques. Algorithm~\ref{alg:offline_synthesis} summarizes this offline synthesis procedure.

\begin{algorithm}[t]
\caption{\CSymPlan{} offline synthesis}
\label{alg:offline_synthesis}
\begin{algorithmic}[1]
\Require Workspace bounds $X_{\mathrm{ws}}$, obstacles $\Obs$, safety margin $d_{\mathrm{safe}}$, velocity bounds $V$
\Require Sampling time $s$, disturbance and measurement bounds $w,z$, grid parameters, input grid $U_{\mathrm{abs}}\subseteq\U$
\Require Task specification: initial set $X_0$, goals $\{X_{\mathrm{goal}}^{(j)}\}_{j=1}^{p}$, with $p=1$ for a single goal
\Ensure Policies $\{\pi^{(j)}\}_{j=1}^{p}$
\State Compute $X_{\mathrm{safe}}$ using \eqref{eq:safe_set}
\State Discretize $X=X_{\mathrm{ws}}\times V$ into uniform cells to obtain $X_{\mathrm{abs}}$
\State Compute $\mathcal{F}_{\mathrm{abs}}$ by over-approximating reachable sets of \eqref{eq:sp_mani_sys}
\State Build the quantizer $Q:X\to X_{\mathrm{abs}}$
\For{$j=1$ to $p$}
    \State Construct abstract sets $\hat X_0^{(j)}$, $\hat X_{\mathrm{safe}}^{(j)}$, and $\hat X_{\mathrm{goal}}^{(j)}$
    \State Synthesize reach-avoid controller $\Pi^{(j)}:X_{\mathrm{abs}}\rightrightarrows U_{\mathrm{abs}}$ by fixed-point computation
    \State Export a selector $\pi^{(j)}(\hat\xi)\in\Pi^{(j)}(\hat\xi)$ as a BDD or lookup table
    \State $X_0^{(j+1)}\gets X_{\mathrm{goal}}^{(j)}$ if $j<p$
\EndFor
\end{algorithmic}
\end{algorithm}




\subsubsection{Real-time policy execution}
\label{sec:offline_policy_exec}

At runtime, the offline-synthesized symbolic policy is executed in a high-frequency control loop. At servo tick $k$, the robot provides joint feedback $(q_k,\dot q_k)$. The task-space state is computed as
\begin{equation*}
    \xi_k=\begin{bmatrix}x_k\\v_k\end{bmatrix}
    =\begin{bmatrix}h(q_k)\\J(q_k)\dot q_k\end{bmatrix}.
\end{equation*}
The quantizer maps $\xi_k$ to an abstract state $\hat\xi_k=Q(\xi_k)$, and the active symbolic policy returns a virtual acceleration $u_k=\pi(\hat\xi_k)$.

The abstract acceleration is realized through the operational-space wrench command in \eqref{eq:total_force},
\begin{equation}
    F_k=\eta(q_k,\dot q_k)+\Lambda(q_k)u_k,
    \label{eq:total_force}
\end{equation}
and the joint torque command
\begin{equation}
    \tau_k=J(q_k)^\top F_k+\tau_{0,k}.
    \label{eq:torque_realization}
\end{equation}
The additional term $\tau_{0,k}$ handles secondary objectives such as posture regulation, damping, and joint-limit avoidance. In our implementation, it is a fixed null-space PD term,
\begin{equation}
    \tau_{0,k}=N(q_k)^\top\left(K_p(q_{\mathrm{ns}}-q_k)-K_d\dot q_k\right),
    \label{eq:tau0_pd_nullspace}
\end{equation}
where $q_{\mathrm{ns}}$ is a nominal posture, $N(q_k)$ is a null-space projector, and $K_p,K_d\succ0$ are fixed gains. These gains are tuned before the experiments and kept constant across trials. Algorithm~\ref{alg:offline_policy_execution} gives the offline runtime loop.

\begin{algorithm}[t]
\caption{\CSymPlan{} offline-policy runtime execution}
\label{alg:offline_policy_execution}
\begin{algorithmic}[1]
\Require Joint feedback $(q,\dot q)$; policies $\{\pi^{(j)}\}_{j=1}^{p}$;
         quantizer $Q$; goal sets $\{X_{\mathrm{goal}}^{(j)}\}_{j=1}^{p}$
\Ensure Torque commands $\tau_k$ at the robot control rate

\State Set active task index $j \gets 1$

\While{robot is running and $j \le p$}
    \State Read $(q_k,\dot q_k)$
    \State Compute $\xi_k \gets
        \bigl[h(q_k)^\top,\,
        \bigl(J(q_k)\dot q_k\bigr)^\top\bigr]^\top$
    \State Quantize $\hat{\xi}_k \gets Q(\xi_k)$
    \State Lookup symbolic input $u_k \gets \pi^{(j)}(\hat{\xi}_k)$
    \State Compute $F_k \gets \eta(q_k,\dot q_k) + \Lambda(q_k)u_k$
    \State Compute $\tau_k \gets J(q_k)^\top F_k + \tau_{0,k}$
    \State Apply $\tau_k$ subject to torque and safety limits

    \If{$\xi_k \in X_{\mathrm{goal}}^{(j)}$ for $N_{\mathrm{hold}}$ consecutive ticks}
        \State $j \gets j + 1$
    \EndIf
\EndWhile
\end{algorithmic}
\end{algorithm}

\subsubsection{Offline guarantee boundary}
\label{sec:offline_guarantee_boundary}

The guarantee provided by \CSymPlan{} is conditional on the abstraction and refinement assumptions, which we state explicitly so that the boundary of the certificate is unambiguous.

\begin{assumption}[Soundness of abstraction and refinement]\label{as:soundness}

(i) The abstract transition relation $\mathcal{F}_{\mathrm{abs}}$ over-approximates all sampled perturbed task-space successors of \eqref{eq:sp_mani_sys}; (ii) the abstract safe and goal sets are linear (conservative) approximations of the continuous sets \eqref{eq:safe_set} and \eqref{eq:goal_set}; (iii) the difference between the intended virtual acceleration and the acceleration realized by \eqref{eq:torque_realization} remains within the disturbance bound $w$ used during synthesis; (iv) the workspace and obstacle model match the physical environment after the configured safety inflation; and (v) the margin $d_{\mathrm{safe}}$ covers the \ee{} and tool geometry, since the safe set is stated for the translational \ee{} position.

\end{assumption} 
\begin{proposition}[Guarantee transfer]\label{prop:transfer}
Under Assumption~\ref{as:soundness}, if the quantized initial state lies in the winning set computed by Algorithm~\ref{alg:offline_synthesis}, then every closed-loop execution of Algorithm~\ref{alg:offline_policy_execution} satisfies the reach-avoid specification \eqref{eq:reach_avoid} with respect to the sets \eqref{eq:safe_set}--\eqref{eq:goal_set}. This follows from the feedback-refinement construction of Section~\ref{sec:symbolic_synthesis}: conditions (i) and (iii) make the quantizer graph $R_Q$ a feedback refinement relation from the sampled perturbed system to $S_{\mathrm{abs}}$, so the refined controller inherits the abstract reach-avoid guarantee \citep{symbolic_feedback,rungger_scots_2016}, while conditions (ii), (iv), and (v) ensure that satisfaction of the abstract specification implies satisfaction of the continuous one.
\end{proposition}
Hardware experiments therefore demonstrate real-time executability together with empirical evidence that the calibrated bounds are conservative enough for Assumption~\ref{as:soundness} to hold in the reported setup.

\begin{remark}[Joint-space instantiation]
The same symbolic-control principle can be instantiated in joint space by defining the abstraction over joint positions and velocities and synthesizing discrete joint-space commands. Such an abstraction would provide joint-level correctness guarantees but would be substantially higher-dimensional for a 7-DOF manipulator. The task-space formulation used in \CSymPlan{} is chosen to make synthesis tractable for reach-avoid manipulation tasks defined by \ee{} position.
\end{remark}

\subsection{Online implementation}
\label{sec:online_implementation}

\begin{figure*}[t]
    \centering
    \includegraphics[width=0.98\textwidth]{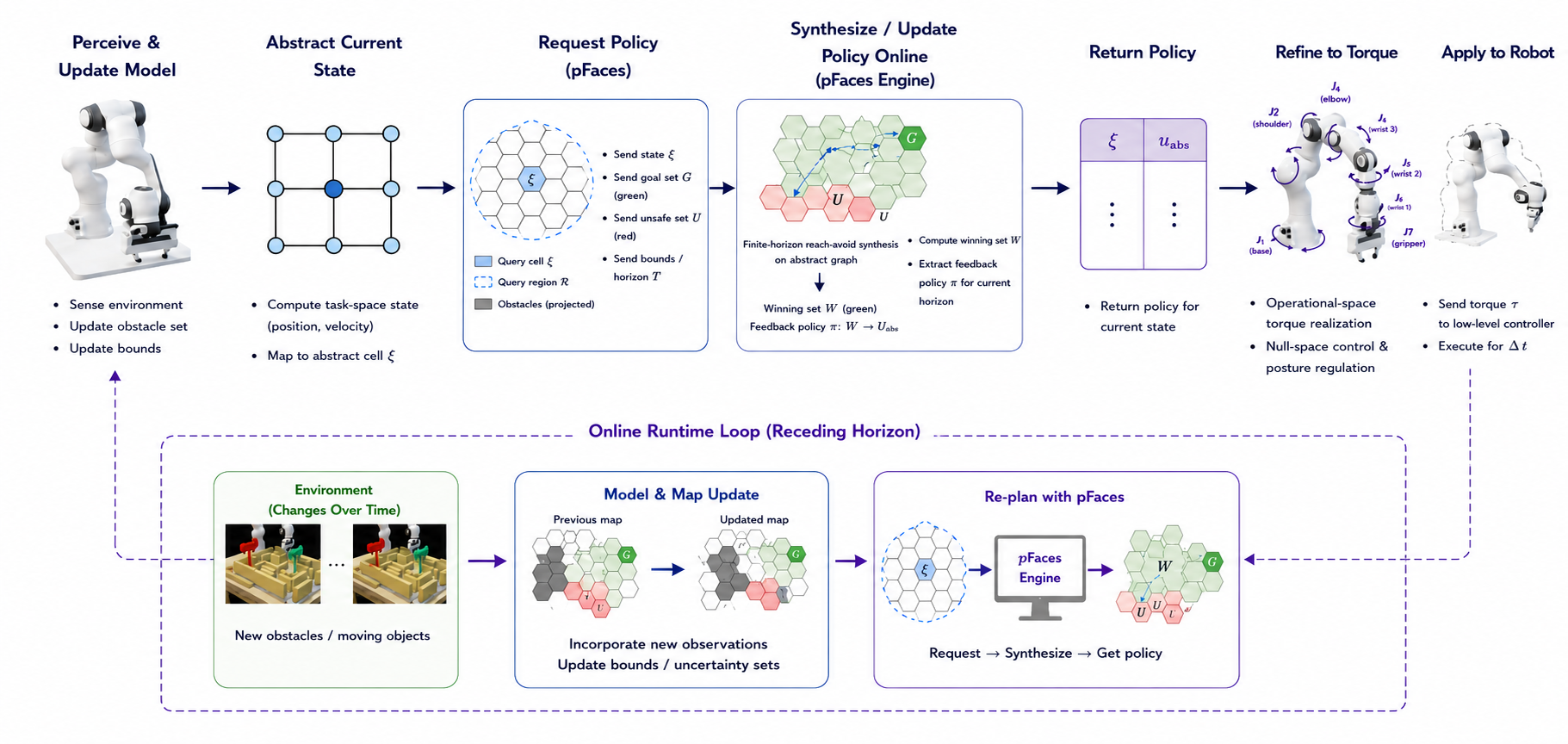}
    \caption{Online \CSymPlan{} architecture. At each replanning instant, the robot senses the environment, updates the obstacle model and uncertainty bounds, maps the measured task-space state to an abstract cell, and sends the current state, target set, unsafe set, and horizon to the pFaces server. pFaces performs finite-horizon reach-avoid synthesis over the symbolic abstraction, returns a certified policy or command segment for the current state, and the robot refines the selected abstract input through operational-space torque realization. The lower loop illustrates the receding-horizon update: changes in the environment trigger map updates, new pFaces requests, and replacement of stale command segments.}
    \label{fig:online_architecture}
\end{figure*}

The online implementation keeps the same sampled perturbed task-space model \eqref{eq:sp_mani_sys}, but moves the symbolic synthesis call into the robot runtime. Instead of exporting a policy once for a fixed scene, an online server maintains the symbolic grid, input grid, sampled post operator, and growth-bound computation. A separate robot process sends the current task data to the server, requests a finite-horizon reach-avoid controller, receives one or more admissible symbolic accelerations, and executes the resulting command segment through the same operational-space torque layer used by the offline implementation. Figure~\ref{fig:online_architecture} shows the resulting architecture and its receding-horizon request--synthesis--execution loop, where the online server calls are implemented through pFaces ~\citep{pfaces}, a tool enabling parallelized computation of reach-avoids synthesis sets as well as the corresponding policies.

\subsubsection{Online task and set representation}
\label{sec:online_sets}

At replanning instant $r$, the online controller receives the measured task-space state in \eqref{eq:online_state},
\begin{equation}
    \xi_r=\begin{bmatrix}h(q_r)\\J(q_r)\dot q_r\end{bmatrix},
    \label{eq:online_state}
\end{equation}
a target region $G_r^x\subset X_{\mathrm{ws}}$, and a runtime obstacle set $\Obs_r\subset X_{\mathrm{ws}}$ from the current task or perception module. The position-level safe set is updated as
\begin{equation}
    X_{\mathrm{safe},r}^x
    =X_{\mathrm{ws}}\setminus
    \set{x\in\R^d\mid \dist(x,\Obs_r)\le d_{\mathrm{safe}}},
    \label{eq:online_safe_position}
\end{equation}
and lifted to the state space by
\begin{equation}
    X_{\mathrm{safe},r}
    =\set{\xi=[x^\top,v^\top]^\top\mid x\in X_{\mathrm{safe},r}^x,\ v\in V}.
    \label{eq:online_safe_state}
\end{equation}
Equations~\eqref{eq:online_safe_position}--\eqref{eq:online_safe_state} define the online safe set. The online goal set is represented as a target tube
\begin{equation}
    X_{\mathrm{goal},r}
    =\set{\xi=[x^\top,v^\top]^\top\mid
    x\in G_r^x,\ \norm{v}_{\infty}\le v_{\mathrm{tol},r}},
    \label{eq:online_goal_state}
\end{equation}
where $v_{\mathrm{tol},r}$ may be chosen strictly for stop-at-goal tasks or relaxed for intermediate waypoints. These sets are received by the online server as interval unions. 
For a collection of axis-aligned boxes $\mathcal{B}=\{[\ell_i,u_i]\}_{i=1}^{N_B}$, the serialization used by the online server is
\begin{equation}
    \mathcal{I}(\mathcal{B})
    =\bigcup_{i=1}^{N_B}
    [\ell_{i,1},u_{i,1}]\times\cdots\times[\ell_{i,2d},u_{i,2d}],
    \label{eq:pfaces_intervals}
\end{equation}
with obstacle boxes written to \texttt{obst\_set}, target boxes written to \texttt{target\_set}, and the measured state written to \texttt{current\_state}. Position-only obstacles are lifted by assigning the velocity coordinates the full bounds $V$.

Let $\eta_x,\eta_u$ denote the state and input grid widths. The online abstraction uses the same finite state and input grids as the offline implementation,
\begin{equation*}
    X_{\mathrm{abs}}=[X_{\mathrm{ws}}\times V]_{\eta_x},
    \qquad
    U_{\mathrm{abs}}=[\U]_{\eta_u}.
\end{equation*}
Only the symbolic safe and target subsets change across replanning calls:
\begin{align}
    \hat X_{\mathrm{safe},r}
    &=\set{\hat\xi\in X_{\mathrm{abs}}\mid
    \mathrm{cell}(\hat\xi)\subseteq X_{\mathrm{safe},r}},\\
    \hat X_{\mathrm{goal},r}
    &=\set{\hat\xi\in X_{\mathrm{abs}}\mid
    \mathrm{cell}(\hat\xi)\subseteq X_{\mathrm{goal},r}}.
    \label{eq:online_abstract_sets}
\end{align}
Thus the abstract sets in \eqref{eq:online_abstract_sets} allow the online implementation to avoid rebuilding the grid at every servo tick; it recomputes the winning set for the current symbolic safe and target sets.

\subsubsection{Online reach-avoid synthesis}
\label{sec:online_pfaces_synthesis}

For each online request, 
a finite-horizon reach-avoid fixed point is computed over the abstract transition relation $\mathcal{F}_{\mathrm{abs}}$. Given horizon $H$, the terminal set is initialized as
\begin{equation*}
    W_{r,H} = \hat X_{\mathrm{goal},r}.
\end{equation*}

For $i = H-1,\ldots,0$, the predecessor update is
\begin{equation}
\begin{aligned}
    W_{r,i}
    ={}& \hat X_{\mathrm{goal},r}
    \cup
    \Bigl\{
        \hat\xi \in \hat X_{\mathrm{safe},r}
        \;\Bigm|\;                                      \\
    &\qquad
        \exists u \in U_{\mathrm{abs}}
        \text{ such that }                              \\
    &\qquad
        \mathcal{F}_{\mathrm{abs}}(\hat\xi,u)
        \subseteq W_{r,i+1}
    \Bigr\}.
\end{aligned}
\label{eq:online_fixed_point}
\end{equation}

The associated time-indexed symbolic controller is the set-valued map
\begin{equation}
\begin{aligned}
    \Pi_{r,i}(\hat\xi)
    =
    \Bigl\{
        u \in U_{\mathrm{abs}}
        \;\Bigm|\;                                      \\
    \qquad
        \mathcal{F}_{\mathrm{abs}}(\hat\xi,u)
        \subseteq W_{r,i+1}
    \Bigr\},
    \qquad
    \hat\xi \in W_{r,i}.
\end{aligned}
\label{eq:online_policy}
\end{equation}

The condition in \eqref{eq:online_policy} is the online analogue of the offline controller lookup: every admissible action keeps all abstract successors inside the remaining winning set. If $Q(\xi_r)\notin W_{r,0}$, the online engine returns no certified action for the requested sets and horizon. The implementation then holds the robot at the last certified reference and may submit a modified request, for example by widening the target tube within configured limits.

When several admissible actions are returned, the controller selects one action for execution. Let
\begin{equation*}
    A_s=
    \begin{bmatrix}
I_d&sI_d\\ 0&I_d
    \end{bmatrix},
\qquad B_s=
    \begin{bmatrix}
\frac{1}{2}s^2I_d\\ sI_d
    \end{bmatrix}
\end{equation*}
be the nominal sampled double-integrator matrices. The implementation ranks candidate actions by a one-step cost
\begin{equation}
\begin{aligned}
    \ell_r(\hat\xi,u)
    &= \alpha_p\norm{x_{T,r}-P_x(A_s\hat\xi+B_su)}_2^2 \\
    &\quad + \alpha_v\norm{P_v(A_s\hat\xi+B_su)}_2^2 \\
    &\quad + \alpha_u\norm{u}_2^2 \\
    &\quad - \alpha_g\Delta_r(\hat\xi,u),
\end{aligned}
    \label{eq:online_action_selection}
\end{equation}

where $P_x$ and $P_v$ project onto position and velocity, $x_{T,r}$ is the target center, and

\begin{equation}
\begin{aligned}
    \Delta_r(\hat\xi,u)
    ={}&
    \norm{x_{T,r}-P_x\hat\xi}_2                                      \\
    &-
    \norm{x_{T,r}-P_x\bigl(A_s\hat\xi+B_su\bigr)}_2 .
\end{aligned}
\label{eq:predicted_progress}
\end{equation}

The term \eqref{eq:predicted_progress} rewards predicted progress of the nominal successor toward the target. Since the minimization is over $\Pi_{r,i}(\hat\xi)$, the heuristic only chooses among actions already certified by the symbolic fixed point. Algorithm~\ref{alg:online_pfaces_synthesis} summarizes one online pFaces synthesis request.

\begin{algorithm}[t]
\caption{\CSymPlan{} online pFaces synthesis request}
\label{alg:online_pfaces_synthesis}
\begin{algorithmic}[1]
\Require Measured state $\xi_r$, target tube $X_{\mathrm{goal},r}$, obstacle set $\Obs_r$, horizon $H$
\Require Fixed grids $X_{\mathrm{abs}}$, $U_{\mathrm{abs}}$, transition relation $\mathcal{F}_{\mathrm{abs}}$, quantizer $Q$
\Ensure Certified command segment $\mathcal{C}_r$ or failure
\State Build $X_{\mathrm{safe},r}$ and $X_{\mathrm{goal},r}$ using \eqref{eq:online_safe_state}--\eqref{eq:online_goal_state}
\State Serialize target and obstacle intervals to pFaces using \eqref{eq:pfaces_intervals}
\State Compute conservative abstract sets $\hat X_{\mathrm{safe},r}$ and $\hat X_{\mathrm{goal},r}$
\State Compute $W_{r,H}\gets \hat X_{\mathrm{goal},r}$
\For{$i=H-1$ down to $0$}
    \State Compute $W_{r,i}$ by the predecessor update \eqref{eq:online_fixed_point}
    \State Store admissible actions $\Pi_{r,i}$ using \eqref{eq:online_policy}
\EndFor
\State $\hat\xi_{r,0}\gets Q(\xi_r)$
\If{$\hat\xi_{r,0}\notin W_{r,0}$}
    \State \Return failure
\EndIf
\For{$i=0$ to $H-1$}
    \State Select $u_{r,i}\in\Pi_{r,i}(\hat\xi_{r,i})$ using \eqref{eq:online_action_selection}
    \State Predict $\hat\xi_{r,i+1}\gets Q(A_s\hat\xi_{r,i}+B_su_{r,i})$
    \State Append $(\hat\xi_{r,i},u_{r,i},\hat\xi_{r,i+1})$ to $\mathcal{C}_r$
    \If{$\hat\xi_{r,i+1}\in\hat X_{\mathrm{goal},r}$}
        \State \textbf{break}
    \EndIf
\EndFor
\State \Return $\mathcal{C}_r$
\end{algorithmic}
\end{algorithm}

\subsubsection{Online buffered execution}
\label{sec:online_buffered_execution}

The robot process separates online synthesis from real-time torque control. A background planning thread communicates with the online server at a lower rate $f_{\mathrm{plan}}$, while the Franka torque callback runs at the servo period. The planner submits a new request when the remaining certified buffer time falls below a threshold $T_{\mathrm{buf}}$:
\begin{equation}
    |\mathcal{C}_{\mathrm{buf}}|\,s\le T_{\mathrm{buf}}.
    \label{eq:replan_trigger}
\end{equation}
Each buffered symbolic action $u_{r,i}$ is interpreted as a constant acceleration over one symbolic sampling interval using the reference in \eqref{eq:online_command_sampling}. For local time $t\in[0,s]$ inside that interval, the desired task-space reference is
\begin{align}
    x_d(t)&=x_{r,i}+t v_{r,i}+\frac{1}{2}t^2u_{r,i},\\
    v_d(t)&=v_{r,i}+t u_{r,i}.
    \label{eq:online_command_sampling}
\end{align}
At servo tick $k$, the online implementation forms the realized acceleration command
\begin{equation}
    u^{\mathrm{rt}}_k
    =
    \mathrm{sat}_{\U}
    \left(
    u_{r,i}+K_x(x_d(t_k)-x_k)+D_x(v_d(t_k)-v_k)
    \right),
    \label{eq:online_rt_acc}
\end{equation}
and uses the same operational-space torque refinement as the offline controller,
\begin{align}
    F_k&=\eta(q_k,\dot q_k)+\Lambda(q_k)u^{\mathrm{rt}}_k,\\
    \tau_k&=J(q_k)^\top F_k+\tau_{0,k}.
    \label{eq:online_torque_realization}
\end{align}
This makes the online synthesis server the decision layer, while the Cartesian or joint impedance behavior remains the torque-realization layer.

Before activating a buffered command, the runtime checks that the current measured state is still close to the state for which the command was certified:
\begin{equation}
    \norm{x_k-x_{r,i}}_2\le \varepsilon_x,
    \qquad
    \norm{v_k-v_{r,i}}_2\le \varepsilon_v.
    \label{eq:stale_command_check}
\end{equation}
Commands that fail \eqref{eq:stale_command_check} are discarded and a new online request is issued. If the buffer is empty, the controller holds the last certified reference using the same impedance layer while waiting for the online synthesis server to generate a new winning set and corresponding symbolic actions. An optional proportional--derivative fallback can be enabled for debugging, but actions generated by that fallback are not part of the certified online controller. Algorithm~\ref{alg:online_runtime_execution} describes the buffered execution loop.

\begin{algorithm}[t]
\caption{\CSymPlan{} online-policy execution}
\label{alg:online_runtime_execution}
\begin{algorithmic}[1]
\Require pFaces, $Q$, targets $\{G_r^x\}$, obstacles $\Obs_r$, $T_{\mathrm{buf}}$
\Ensure Torque commands $\tau_k$

\State $r\gets 1$, \quad $\mathcal C_{\mathrm{buf}}\gets\emptyset$, \quad $c_{\mathrm{act}}\gets\emptyset$
\State Start asynchronous online worker (via pFaces)

\While{robot is running}
    \State $\xi_k \gets \mathrm{Sense}(q_k,\dot q_k)$

    \If{$\mathrm{LowBuffer}(\mathcal C_{\mathrm{buf}},T_{\mathrm{buf}})$ and $\mathrm{Idle}()$}
        \State $\mathrm{PlanAsync}(\xi_k,G_r^x,\Obs_r)$
    \EndIf

    \State $\mathcal C_{\mathrm{buf}}
    \gets \mathrm{AppendCertified}(\mathcal C_{\mathrm{buf}})$

    \If{$c_{\mathrm{act}}=\emptyset$}
        \State $c_{\mathrm{act}}
        \gets \mathrm{PopValid}(\mathcal C_{\mathrm{buf}},\xi_k)$
    \EndIf

    \If{$c_{\mathrm{act}}\neq\emptyset$}
        \State $(x_d,v_d)\gets \mathrm{SampleReference}(c_{\mathrm{act}})$
        \State $u_k^{\mathrm{rt}}\gets \mathrm{TrackCertified}(\xi_k,x_d,v_d)$
    \Else
        \State $u_k^{\mathrm{rt}}\gets \mathrm{HoldReference}(\xi_k)$
    \EndIf

    \State $\tau_k\gets \mathrm{RealizeTorque}(u_k^{\mathrm{rt}},q_k,\dot q_k)$
    \State Apply $\tau_k$ with torque, rate, and safety limits

    \If{$\mathrm{GoalReached}(\xi_k,G_r^x)$}
        \State $r\gets r+1$, \quad $c_{\mathrm{act}}\gets\emptyset$, \quad
        $\mathcal C_{\mathrm{buf}}\gets\emptyset$
    \EndIf
\EndWhile
\end{algorithmic}
\end{algorithm}

\subsubsection{Online guarantee boundary}
\label{sec:online_guarantee_boundary}

The online guarantee is the same reach-avoid guarantee as in the offline case, but it is local to the sets, horizon, and state used in each online server request. Formally, Proposition~\ref{prop:transfer} applies per request, with $X_{\mathrm{safe},r}$, $X_{\mathrm{goal},r}$, and the finite horizon $H$ in place of the fixed offline sets. If $Q(\xi_r)\in W_{r,0}$, then any command selected from \eqref{eq:online_policy} keeps the sampled perturbed system inside $X_{\mathrm{safe},r}$ and reaches $X_{\mathrm{goal},r}$ within the requested horizon, provided that the disturbance, measurement, and torque-refinement bounds used by $\mathcal{F}_{\mathrm{abs}}$ contain the physical realization error. The guarantee is preserved across replanning calls when each accepted buffered command satisfies the state-matching condition \eqref{eq:stale_command_check} and the runtime obstacle model remains valid until the next certified update.

The guarantee does not cover three cases: no winning action returned by the online server, execution of an explicitly enabled uncertified fallback, or environment changes that invalidate $X_{\mathrm{safe},r}$ before a new certified request has been accepted. In these cases the implementation either holds the last certified reference, widens the target tube within configured bounds, submits a new request to compute another winning action, or stops the task if the state leaves the symbolic domain.

\section{Experimental Evaluation}
\label{sec:experiments}

The experimental evaluation mirrors the implementation structure. The offline evaluation tests the precomputed-controller pipeline in static workspaces, including real Franka FR3 deployment and PyBullet benchmarking against planning and receding-horizon baselines. The online evaluation tests the request--synthesis--execution loop under changing task geometry, including Isaac Sim/Isaac Lab \citep{isaac_sim} task families and real Franka experiments with perception-driven obstacle updates.

\subsection{Offline evaluation}
\label{sec:offline_evaluation}

The offline evaluation studies the precomputed-policy version of \CSymPlan{} in fixed workspaces. The aim is to verify that the quantization--lookup--torque pipeline can be executed on hardware and to compare the offline symbolic feedback policy against representative planning-and-control baselines in simulation. In contrast to the online implementation in Section~\ref{sec:online_implementation}, the obstacle set, workspace, and goal sequence are fixed during each offline synthesis run, and the resulting policy is exported before execution.

\subsubsection{Offline real-world Franka experiments}
\label{sec:offline_real_franka}

\begin{figure}[t]
    \centering
    \includegraphics[width=0.95\columnwidth]{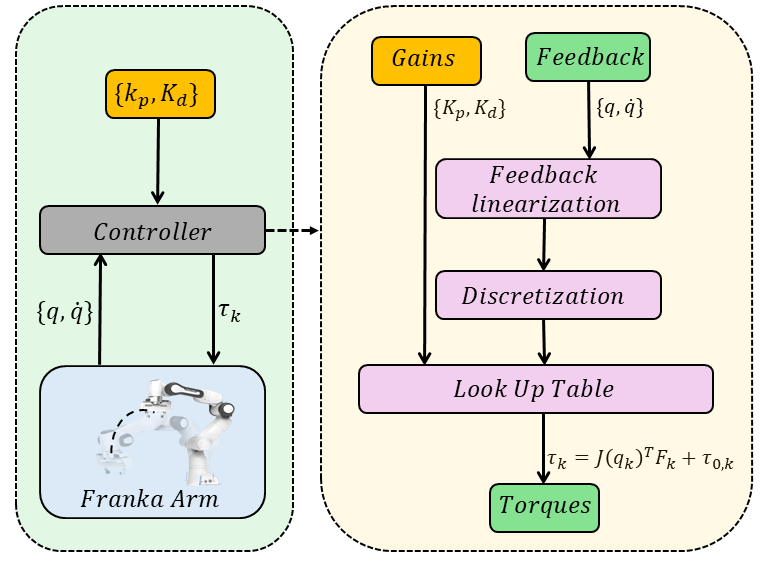}
    \caption{Offline real-time execution architecture on the Franka FR3. Joint feedback is converted to task-space position and velocity, quantized, queried through the precomputed symbolic policy, and refined to joint torques at the control rate.}
    \label{fig:offline_franka_rt_arch}
\end{figure}

\paragraph{Offline hardware and control stack.}
The offline synthesized policy is deployed on a Franka FR3 using the architecture in Figure~\ref{fig:offline_franka_rt_arch}. The system has two stages. The offline stage uses SCOTS~\citep{rungger_scots_2016} to construct the finite abstraction and synthesize a reach-avoid controller for a fixed workspace and task. The resulting controller is exported as a lookup structure. The runtime stage performs state estimation, quantization, policy lookup, and torque realization at the robot control rate. The control loop runs at $1\,\mathrm{kHz}$, and the low-level gains in \eqref{eq:tau0_pd_nullspace} are fixed across all trials.

\begin{table*}[!t]
\centering
\caption{Offline simulated reach-avoid benchmarking in PyBullet across randomized trials and scenes. Metrics summarize reliability, efficiency, control quality, and online computation time. }
\label{tab:offline_sim_results}
\vspace{1mm}
\scriptsize
\setlength{\tabcolsep}{4pt}
\renewcommand{\arraystretch}{1.15}
\resizebox{\textwidth}{!}{%
\begin{tabular}{@{}lccccccc@{}}
\toprule
\textbf{Method} &
\textbf{Success (\%)} &
\textbf{Violation (\%)} &
\textbf{$t_{\mathrm{goal}}$ (s)} &
\textbf{$L$ (m)} &
\textbf{Smoothness} &
\textbf{Effort} &
\textbf{CPU/step (ms)} \\
\midrule
\CSymPlan{}          & \textbf{100} & \textbf{0.0} & \textbf{2.3} & 1.25 & 0.90 & 1.05 & \textbf{0.05 / 0.08} \\
RRT + tracker        & 76 & 10.0 & 3.9 & 1.42 & 1.25 & 1.18 & 120 / 240 \\
LQR-RRT$^\star$ + tracker & 83 & 6.0 & 3.6 & 1.33 & 0.62 & 1.02 & 170 / 340 \\
MPPI                 & 95 & 4.5 & 2.5 & 1.17 & 0.52 & 0.98 & 18.1 / 19.0 \\
Neural MPC           & 93 & 3.5 & 2.6 & \textbf{1.14} & \textbf{0.38} & \textbf{0.92} & 3.6 / 4.4 \\
\bottomrule
\end{tabular}%
}
\vspace{1mm}
\begin{minipage}{0.98\textwidth}
\footnotesize
\textbf{Notes:} $N$ denotes the number of trials and $\Delta t$ the control period. Success $=100N_{\mathrm{succ}}/N$; violation $=100N_{\mathrm{viol}}/N$, where a violation means $\exists k:x_k\notin X^x_{\mathrm{safe}}$; $t_{\mathrm{goal}}=\min\{k\Delta t:\xi_k\in X_{\mathrm{goal}}\}$; $L=\sum_k\norm{x_{k+1}-x_k}_2$; smoothness $=\sum_k\norm{u_{k+1}-u_k}_2^2$; effort $=\sum_k\norm{\tau_k}_2^2$. CPU/step reports the mean and 95th percentile over the execution loop. 
\end{minipage}
\end{table*}

\begin{figure}[t]
    \centering
    \includegraphics[width=0.95\columnwidth]{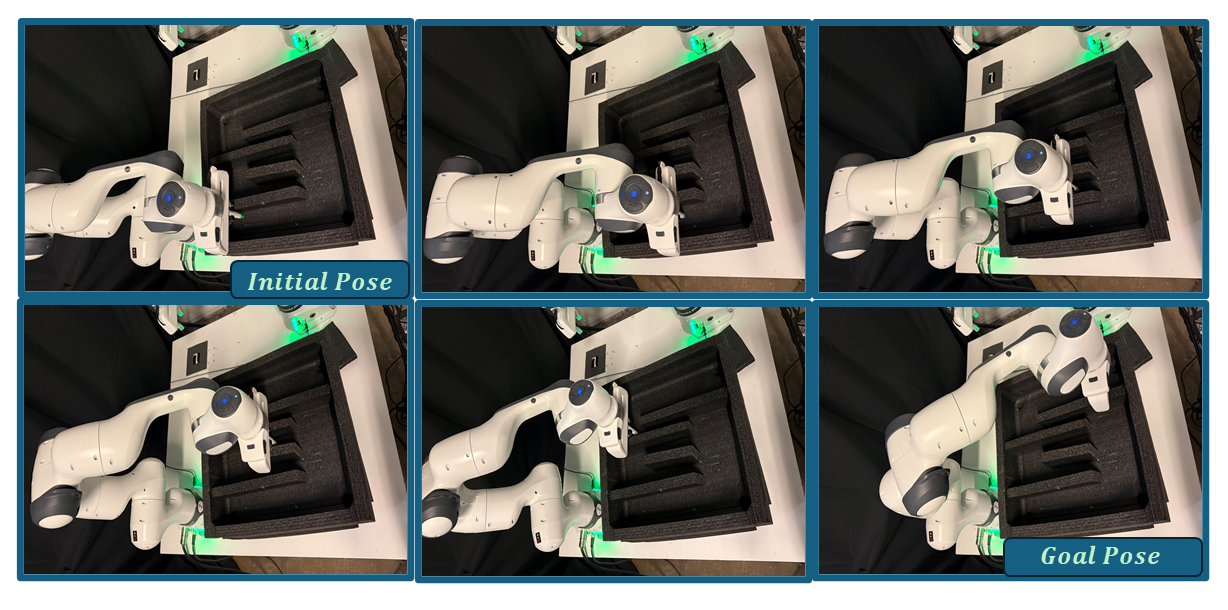}
    \caption{Offline hardware experiment procedure and control-loop execution order.}
    \label{fig:offline_fr3_procedure}
\end{figure}

\paragraph{Offline tasks and procedure.}
The real-robot task is a task-space reach-avoid problem defined by the safe and goal sets in \eqref{eq:safe_set}--\eqref{eq:goal_set}. The initial set is a neighborhood around a nominal start pose. A trial is counted as successful if the \ee{} enters the goal set and remains there for $N_{\mathrm{hold}}$ consecutive control ticks while staying in the safe set for the entire execution. We test five static scenes with increasing obstacle complexity. Obstacles are represented by boxes, cylinders, and mesh primitives, and the same execution stack and actuation limits are used across scenes.

Each hardware trial follows the procedure in Figure~\ref{fig:offline_fr3_procedure}: initialize the robot to the start configuration, verify that the measured state lies in the winning set, execute the $1\,\mathrm{kHz}$ quantization--lookup--torque loop, and terminate on success, timeout, or safety stop. We repeat five trials per scene with fixed initial states and fixed goals.

\begin{figure*}[t!]
    \centering
    \begin{minipage}[t]{0.32\textwidth}
        \centering
        \includegraphics[width=\linewidth]{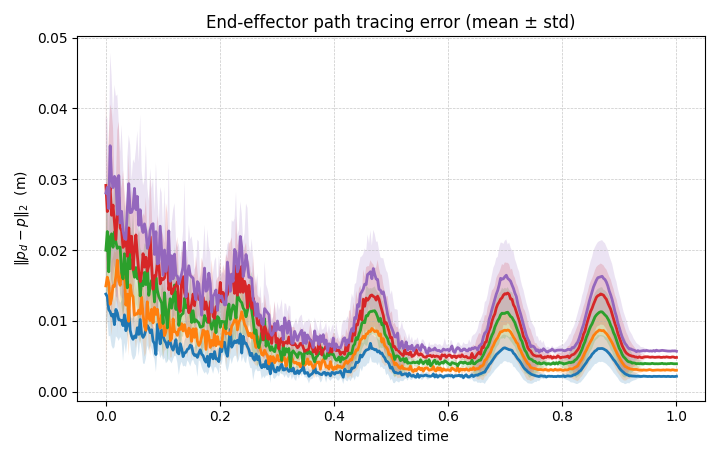}\\[-0.3em]
        {\footnotesize (a) End-effector tracking error}
    \end{minipage}
    \hfill
    \begin{minipage}[t]{0.32\textwidth}
        \centering
        \includegraphics[width=\linewidth]{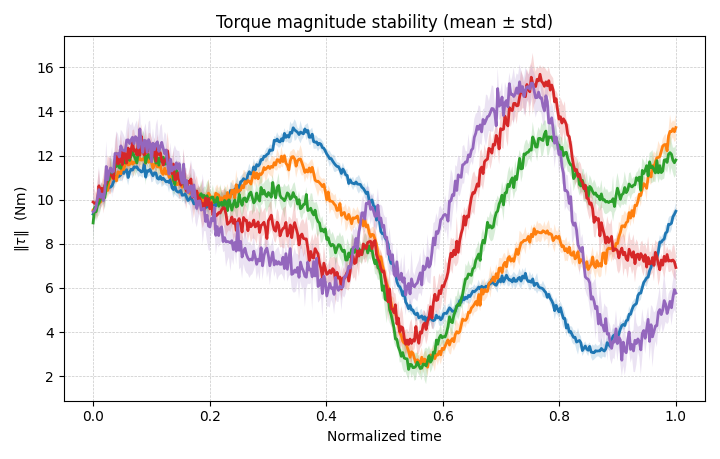}\\[-0.3em]
        {\footnotesize (b) Commanded joint-torque norm}
    \end{minipage}
    \hfill
    \begin{minipage}[t]{0.32\textwidth}
        \centering
        \includegraphics[width=\linewidth]{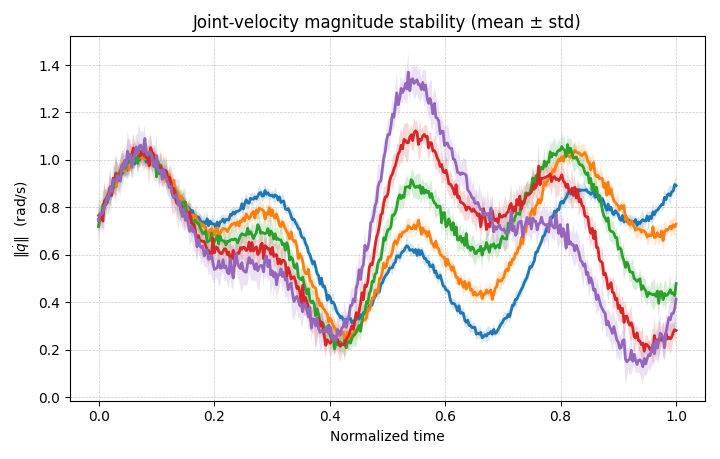}\\[-0.3em]
        {\footnotesize (c) Measured joint-velocity norm}
    \end{minipage}

    \caption{Offline Franka FR3 real-robot execution metrics across hardware trials. 
    The panels show (a) end-effector position tracking error, (b) commanded joint-torque norm, and 
    (c) measured joint-velocity norm. Curves show the mean and one-standard-deviation band over 
    $N=5$ trials per scene.}
    \label{fig:offline_franka_real_robot_metrics}
\end{figure*}

\paragraph{Offline real-world metrics.}
We report aggregate hardware metrics over $N=5$ trials per scene. Curves show the mean and one-standard-deviation band over normalized execution time. Figure~\ref{fig:offline_franka_real_robot_metrics} shows the \ee{} position tracking error,
\begin{equation*}
    e_p(t)=\norm{p_d(t)-p(t)}_2,
\end{equation*}
where $p_d(t)$ is the desired \ee{} position command and $p(t)$ is the measured \ee{} position from forward kinematics. The same figure also reports the commanded joint-torque norm $\norm{\tau(t)}_2$ and the measured joint-velocity norm $\norm{\dot q(t)}_2$.




\subsubsection{Offline simulated benchmarking}
\label{sec:offline_simulated_benchmarking}

\begin{figure}[t]
    \centering
    \includegraphics[width=0.9\columnwidth,trim=1.2cm 0.8cm 1.2cm 0.6cm,clip]{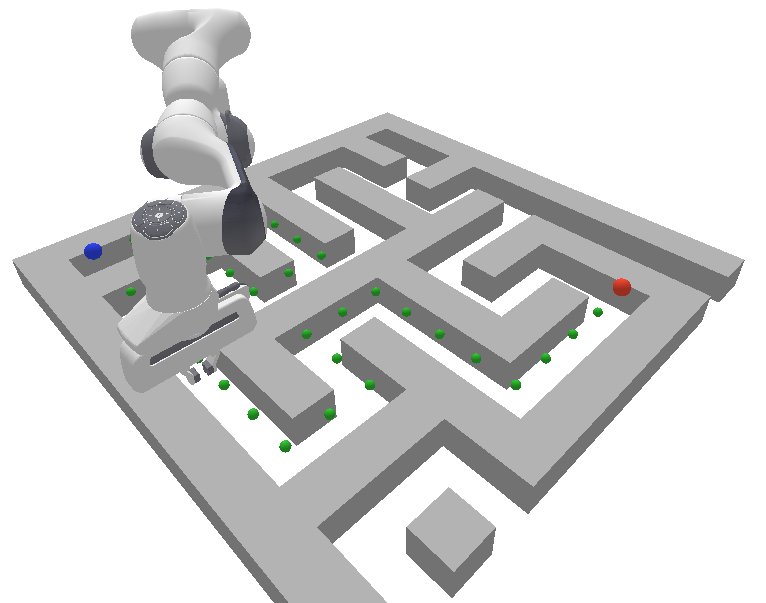}
    \caption{Representative offline simulation scene. The robot starts at the blue marker and aims to reach the red marker. Green markers indicate a trajectory produced by \CSymPlan{} through the certified safe set.}
    \label{fig:offline_maze_scene}
\end{figure}

We also evaluate the offline version of \CSymPlan{} in simulation against representative planning-and-control baselines. The simulated study uses a Franka-family manipulator model in PyBullet with the same task-space bounds, obstacle definitions, safety margins, actuation limits, and torque-realization interface across all methods. Figure~\ref{fig:offline_maze_scene} shows a representative simulated scene. Each method outputs a task-space command or reference that is executed through the same low-level realization layer. Thus, the comparison isolates the decision layer while keeping collision checking, limits, and dynamics consistent.

\paragraph{Offline task setup.}
We consider five reach-avoid tasks in environments with randomized starts, goals, and obstacle placements sampled from fixed configuration sets. A trial terminates on success, safety violation, or timeout. Success requires reaching $X_{\mathrm{goal}}$ while remaining in $X_{\mathrm{safe}}$ for the entire trajectory. Safety violation is registered if the \ee{} position leaves the position-level safe set, intersects the inflated obstacle set, or violates configured velocity or torque limits. All simulated experiments are run on a workstation with an NVIDIA RTX 5070 Ti GPU with $16\,\mathrm{GB}$ VRAM and $64\,\mathrm{GB}$ RAM.

\paragraph{Offline baselines.}
We compare against four representative baselines for the offline setting:
\begin{itemize}[leftmargin=*]
    \item \textbf{RRT + tracker.} A geometric RRT computes a collision-free path that is executed by the shared tracking and torque-realization layer.
    \item \textbf{LQR-RRT$^\star$ + tracker.} LQR-RRT$^\star$ grows a state-space tree using an LQR-based steering metric and then executes the resulting trajectory through a tracking controller~\citep{adityaLQRRRT}.
    \item \textbf{MPPI.} Model predictive path integral control samples control sequences over a finite horizon, rolls them out through the dynamics model, and applies the first action using path-integral reweighting~\citep{williams_mppi_2017}.
    \item \textbf{Neural MPC.} Neural MPC computes receding-horizon actions under constraints, using a learned approximation to reduce latency while retaining a constraint-aware formulation~\citep{adityaNEURALMPC}.
\end{itemize}

\paragraph{Offline metrics and results.}
Table~\ref{tab:offline_sim_results} summarizes reliability, motion quality, and deployability metrics. Reliability is measured by success and violation rates. Motion quality is measured by time-to-goal, path length, smoothness, and effort. Deployability is measured by mean and 95th-percentile online computation time per control step.

\begin{figure*}[!t]
    \centering
    \begin{minipage}[t]{0.19\textwidth}
        \centering
        \includegraphics[width=\linewidth]{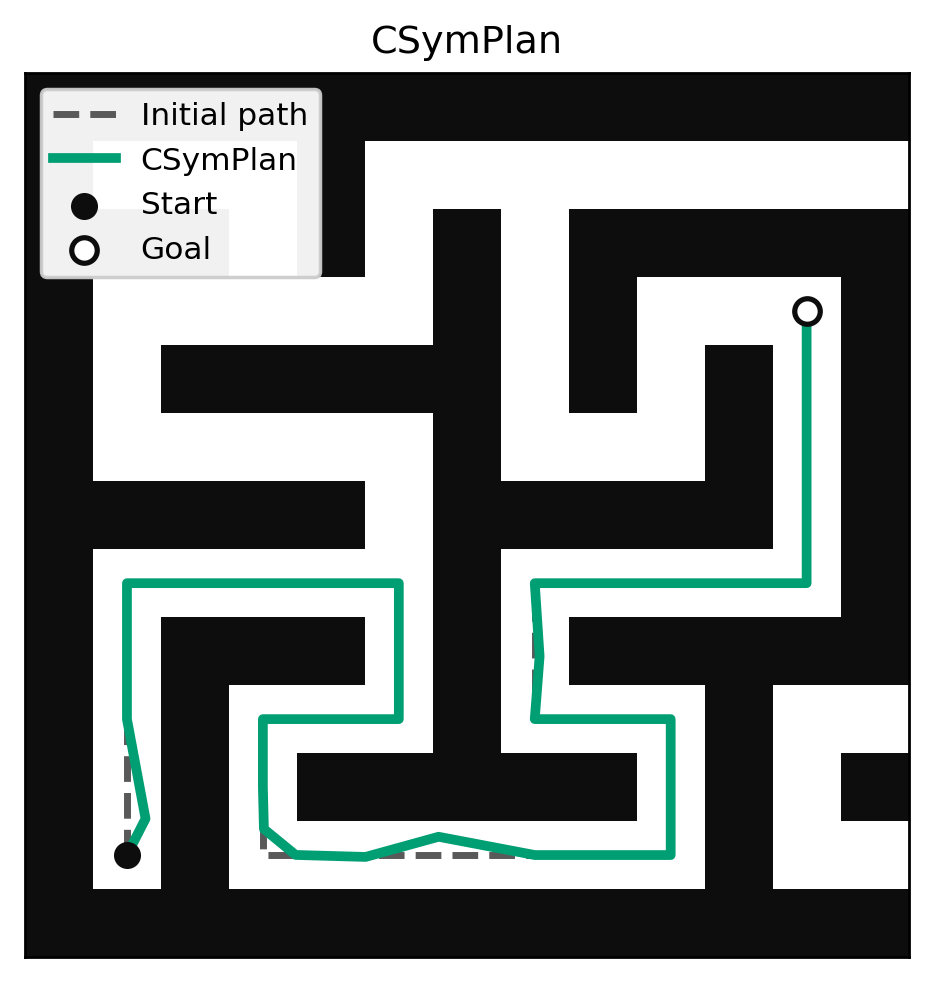}\\[-0.3em]
        {\footnotesize (a) \CSymPlan{}}
    \end{minipage}\hfill
    \begin{minipage}[t]{0.19\textwidth}
        \centering
        \includegraphics[width=\linewidth]{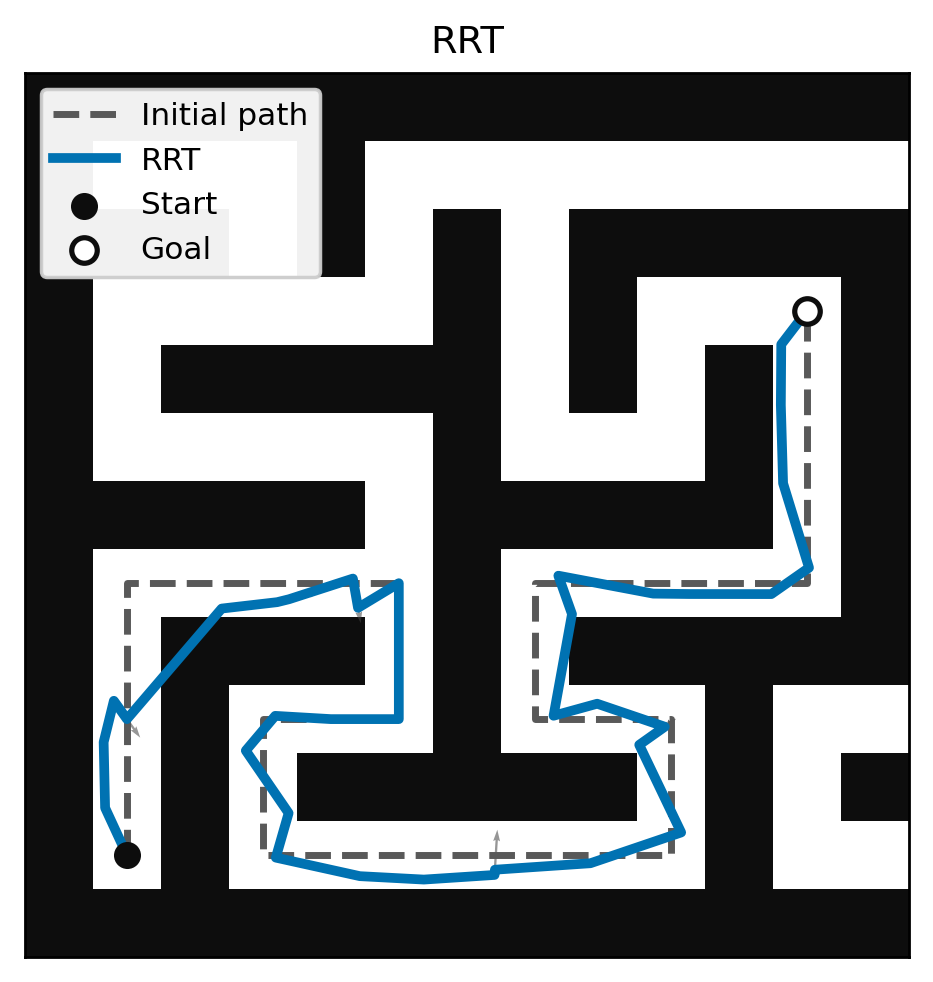}\\[-0.3em]
        {\footnotesize (b) RRT}
    \end{minipage}\hfill
    \begin{minipage}[t]{0.19\textwidth}
        \centering
        \includegraphics[width=\linewidth]{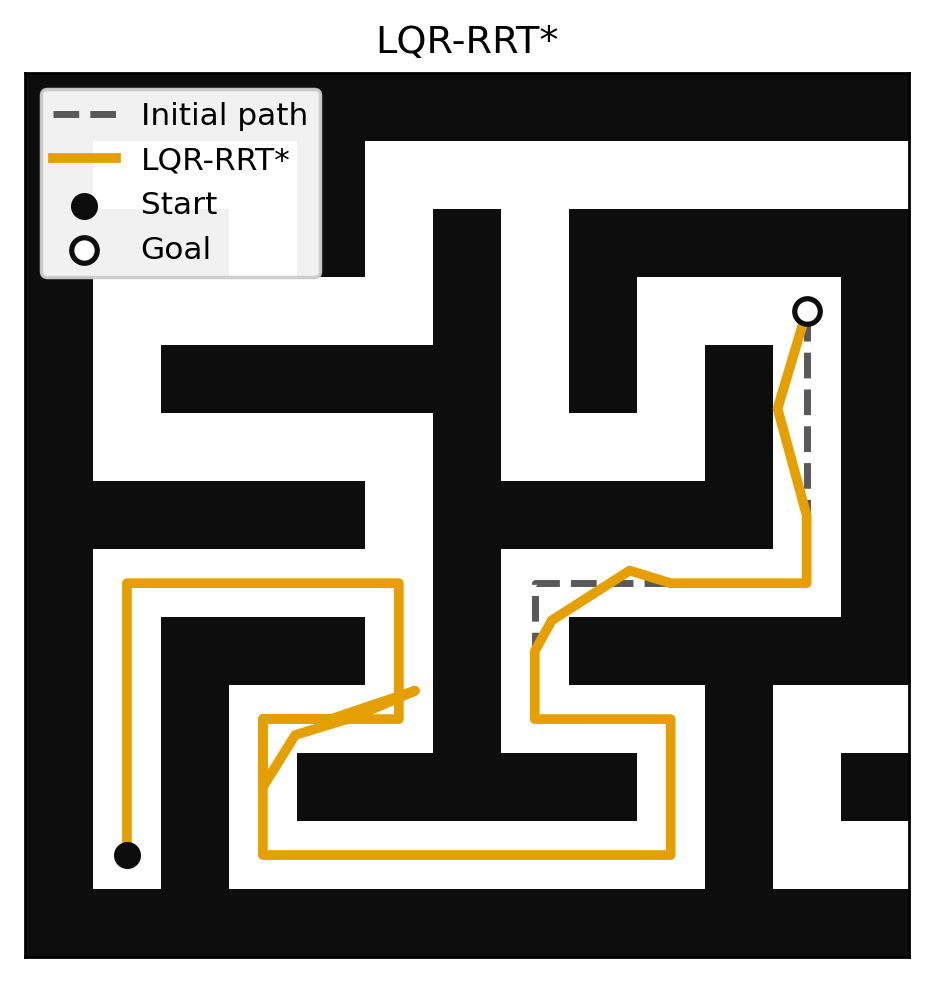}\\[-0.3em]
        {\footnotesize (c) LQR-RRT$^\star$}
    \end{minipage}\hfill
    \begin{minipage}[t]{0.19\textwidth}
        \centering
        \includegraphics[width=\linewidth]{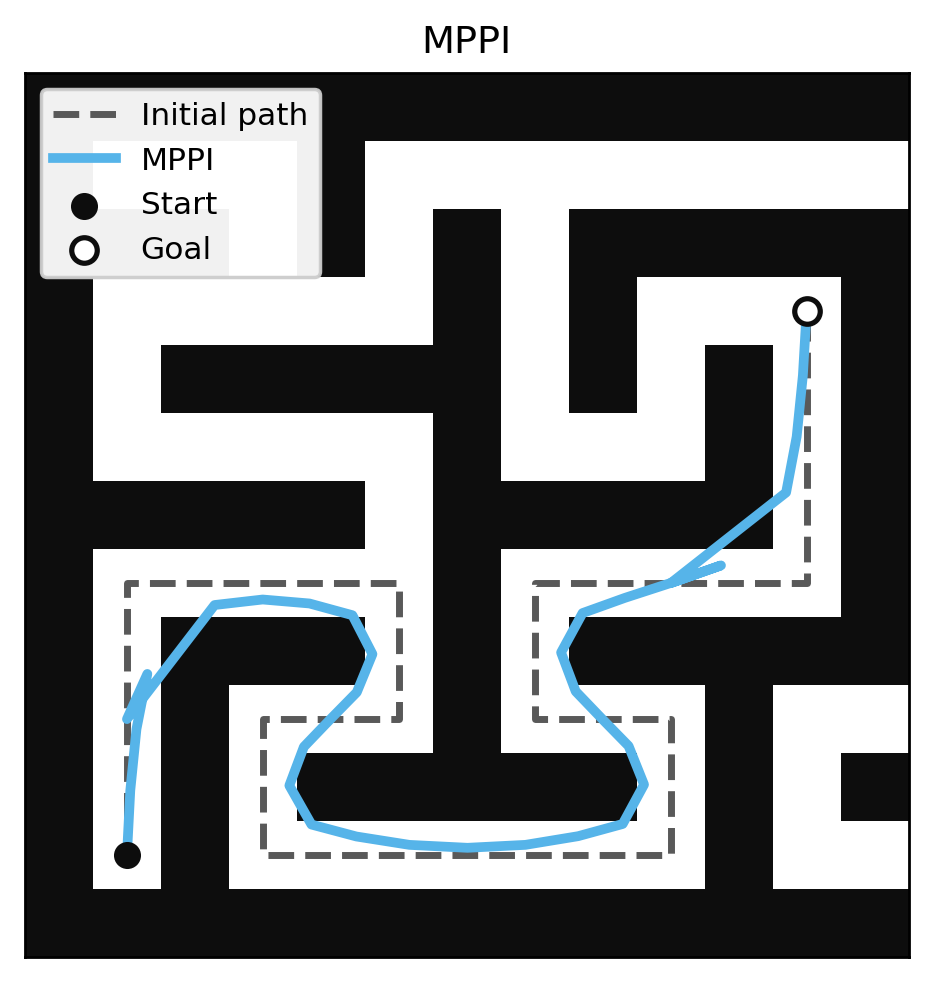}\\[-0.3em]
        {\footnotesize (d) MPPI}
    \end{minipage}\hfill
    \begin{minipage}[t]{0.19\textwidth}
        \centering
        \includegraphics[width=\linewidth]{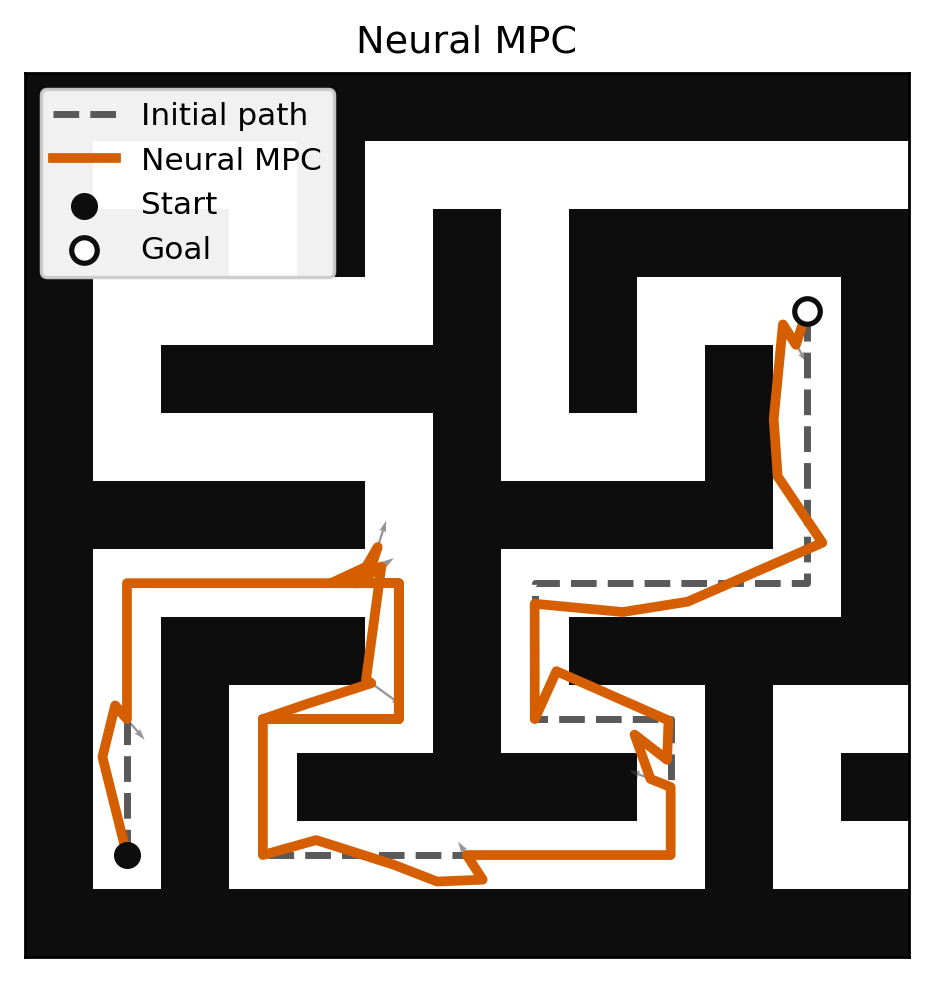}\\[-0.3em]
        {\footnotesize (e) Neural MPC}
    \end{minipage}

    \caption{Representative maze trajectories for offline \CSymPlan{} and baselines. Dashed curves show reference paths where applicable and solid curves show executed \ee{} trajectories from the same start to the same goal. Black regions are obstacles; a collision is counted when the \ee{} footprint intersects an obstacle cell.}
    \label{fig:offline_fr3_baseline_maze_trajs}
\end{figure*}

Across the randomized trials, \CSymPlan{} achieves 100\% success with no recorded safety violations. Figure~\ref{fig:offline_fr3_baseline_maze_trajs} shows representative executed trajectories for \CSymPlan{} and the baselines. The receding-horizon baselines produce competitive time-to-goal and path-length values but have nonzero violation rates in tight scenes. LQR-RRT$^\star$ improves over geometric RRT by incorporating dynamics into the planning stage, but it still inherits failure modes from the plan--then--track separation. The main computational trade-off is offline synthesis: \CSymPlan{} incurs a larger one-time cost to build the abstraction and solve the fixed point, but the resulting runtime policy query is a constant-time lookup.

\noindent\textbf{Offline compute cost.} In the reported simulated setup, \CSymPlan{} required $T_{\mathrm{offline}}=290\,\mathrm{s}$ in total, comprising $T_{\mathrm{abs}}=75\,\mathrm{s}$ for abstraction construction and $T_{\mathrm{syn}}=215\,\mathrm{s}$ for fixed-point synthesis. As a reference, the average offline planning time for RRT was $6.2\,\mathrm{s}$. This difference is expected because \CSymPlan{} performs conservative abstraction construction and fixed-point synthesis, which is substantially heavier than incremental sampling. The benefit is that this cost is paid once per scene and model; after synthesis, online execution is a predictable quantize--lookup--apply operation. When tasks share the same workspace and dynamics, the abstraction phase can be reused and only re-synthesis is required for new goals, partially amortizing the offline burden.

\subsection{Online evaluation}
\label{sec:online_evaluation}
\subsubsection{Online simulated Isaac Sim experiments}
\label{sec:online_simulated_isaac_experiments}

\paragraph{Online simulation stack.}
\label{sec:isaac_stack}

The simulated experiments use the \texttt{CSymIsaac} environment, which runs a Franka-family manipulator in Isaac Sim through Isaac Lab. The symbolic state and input grids and the transition relation $\mathcal{F}_{\mathrm{abs}}$ are constructed once before execution and reused across replanning calls. At runtime, only the current safe and target sets are updated from the robot state and scene geometry.

We use pFaces~\citep{pfaces} as an online synthesis server over this fixed abstraction. At each replanning step, the measured state, target set, obstacle-derived safe set, and horizon are sent to pFaces, which computes the finite-horizon winning set and corresponding admissible symbolic actions. The returned action is interpreted as a Cartesian acceleration, checked for staleness, and executed through the task-space realization layer in \eqref{eq:online_rt_acc}--\eqref{eq:online_torque_realization}. This closes the state/perception--synthesis--realization loop used in both simulation and hardware.

\paragraph{Online simulated task families.}
\label{sec:isaac_tasks}

\begin{figure*}[t]
    \centering
    \includegraphics[width=\textwidth]{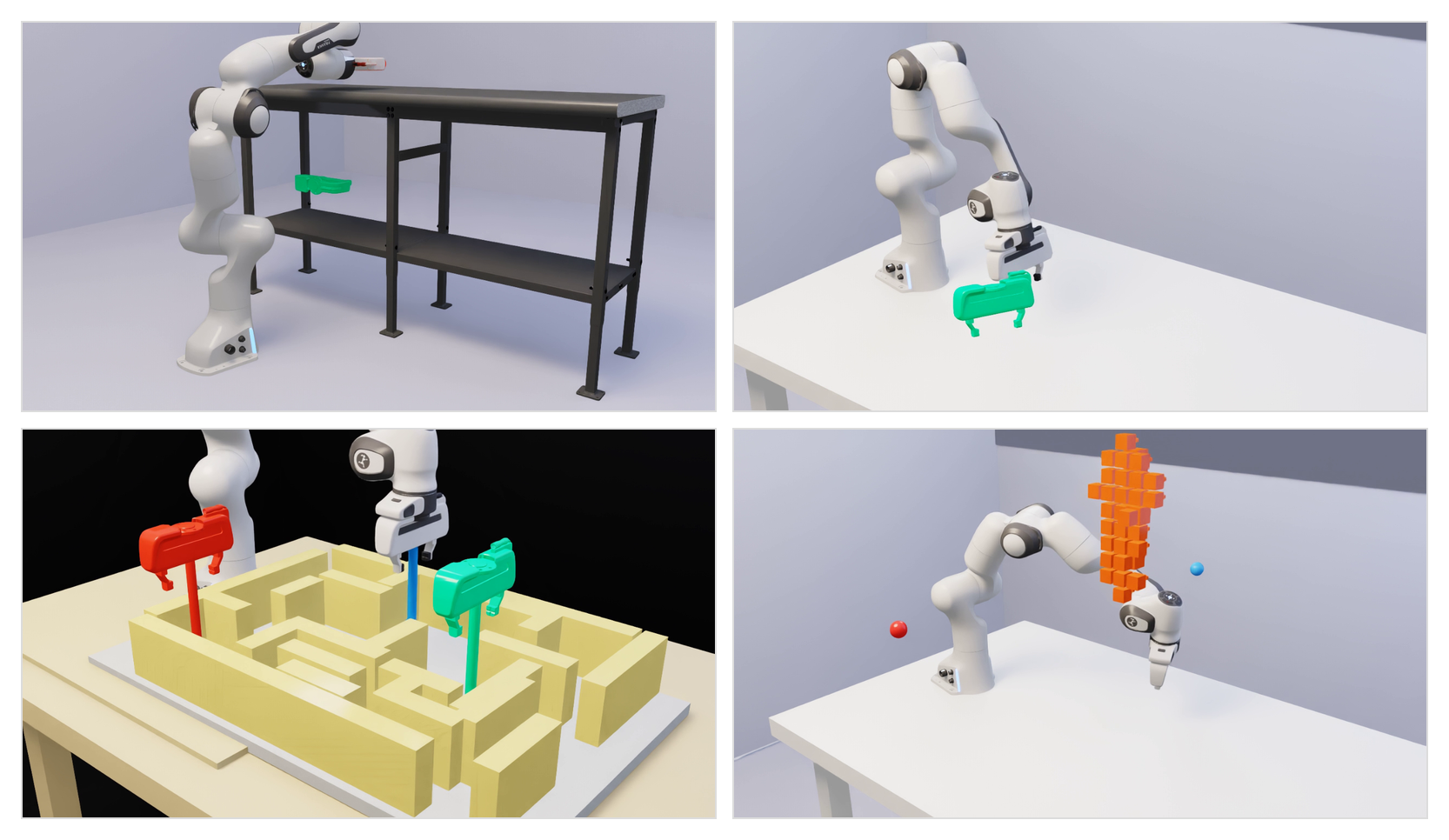}
    \caption{Representative Isaac Sim benchmark environments used for online evaluation: shelf/rack reach-avoid, point reach, 3D maze, and moving-obstacle under tasks.}
    \label{fig:online_sim_benchmark_envs}
\end{figure*}

We evaluate three families of reach-avoid tasks, illustrated in Figure~\ref{fig:online_sim_benchmark_envs}.
\begin{itemize}[leftmargin=*]
\item \textbf{Shelf reach-avoid.} The robot moves between targets placed at different parts of a shelf or rack, including top-shelf approach, shelf insertion, retreat, and lower-shelf placement motions. This task stresses three-dimensional motion because the feasible path must coordinate lateral, depth, and vertical movement while avoiding shelf surfaces, posts, and the shelf lip.
\item \textbf{Three-dimensional maze.} The robot moves a probe-like end-effector through a generated tabletop maze. The safe set is the corridor volume, and the obstacle set is formed by the maze walls and baffles. This task evaluates whether the online symbolic controller can keep the executed trajectory inside a narrow connected safe region rather than simply moving around isolated convex obstacles.
\item \textbf{Moving obstacle reach-avoid.} The robot moves between two Cartesian goal regions while a voxelized cube obstacle changes occupancy over time. The obstacle can sweep laterally, switch between high and low states, or require over-under behavior. This task evaluates online replanning when the safe set changes during execution.
\end{itemize}

\paragraph{Online simulated procedure.}
\label{sec:isaac_procedure}

Each simulated trial begins by resetting the Franka model, loading the task geometry, and selecting a start state, target tube, and obstacle set. For shelf and maze tasks, the target sequence is known at the start of the episode, but the controller still solves each segment online from the current measured state. For moving-obstacle tasks, the obstacle set is updated whenever the voxel state changes, and a new online request via pFaces is issued when the certified command buffer falls below the replanning threshold in \eqref{eq:replan_trigger}. A trial terminates when the final target has been held for the required dwell time, when the \ee{} leaves the safe set, when a robot or scene collision is detected, or when the episode reaches a timeout.

The main simulated measurements are success or failure, safety violation, time-to-goal, path length, minimum obstacle clearance, command smoothness, joint-torque effort, number of pFaces replanning calls, and pFaces synthesis latency. These metrics and benchmark results are summarized in Section~\ref{sec:online_metrics_results}.

\subsubsection{Online real-world Franka experiments}
\label{sec:online_real_franka_experiments}

\paragraph{Online hardware and control stack.}
\label{sec:real_franka_stack}

The hardware experiments use the \texttt{CSymFranka} stack on a Franka FR3. The robot reports joint position and velocity through libfranka at the servo rate, and similar to the simulated case, a pFaces controller takes synthesis requests.
In particular, the controller computes the measured \ee{} state in the Franka base frame, submits online pFaces synthesis requests in a background thread, buffers the returned symbolic accelerations, and realizes the active command as a Cartesian impedance or joint-impedance torque command. The torque layer is kept fixed across real experiments so that differences between tasks are attributed to the online symbolic decision layer and the task geometry rather than to retuning the low-level controller.

\begin{figure*}[!t]
    \centering
    \includegraphics[width=0.98\textwidth]{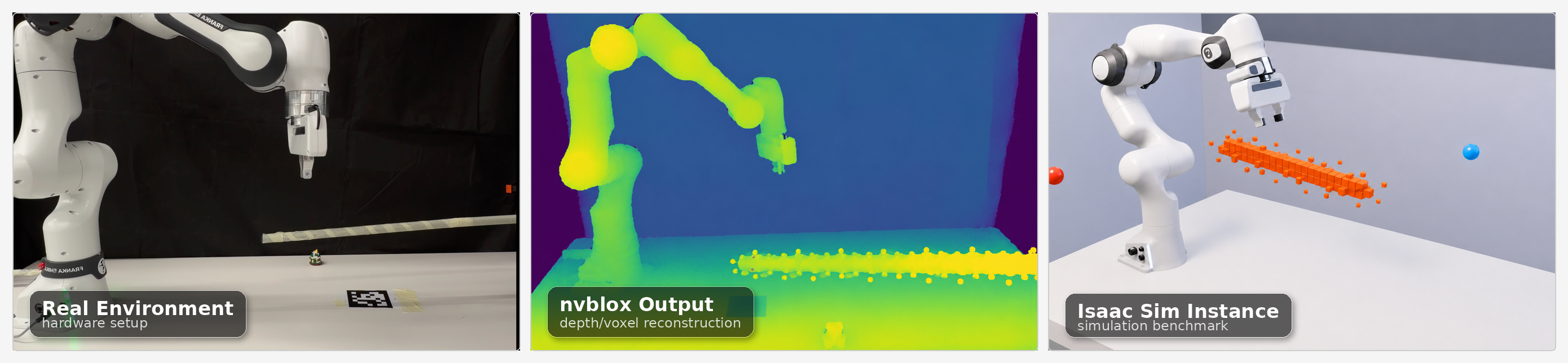}
    \caption{Real-world online perception and simulation correspondence. Left: Franka FR3 hardware setup used for the online reach-avoid experiments. Middle: DNVblox voxel reconstruction obtained from depth sensing and used to form the runtime obstacle set. Right: corresponding Isaac Sim scene used for controlled online benchmarking.}
    \label{fig:online_real_nvblox_vis}
\end{figure*}

Real-time obstacle tracking is provided by DNVblox \cite{nvblox}, which fuses depth observations into a voxel map and publishes the occupied obstacle set used by the online pFaces request; Figure~\ref{fig:online_real_nvblox_vis} shows the perception and simulation correspondence. In the real-robot experiments, the DNVblox obstacle map is updated at $10\,\mathrm{Hz}$, while the libfranka torque callback remains at the robot servo rate. Each pFaces replanning request therefore uses the most recent DNVblox obstacle snapshot, inflated by the configured safety margin $d_{\mathrm{safe}}$ before being serialized into \texttt{obst\_set}.

\paragraph{Online real-world task families.}
\label{sec:real_franka_tasks}

\begin{figure*}[!t]
    \centering
    \includegraphics[width=0.98\textwidth]{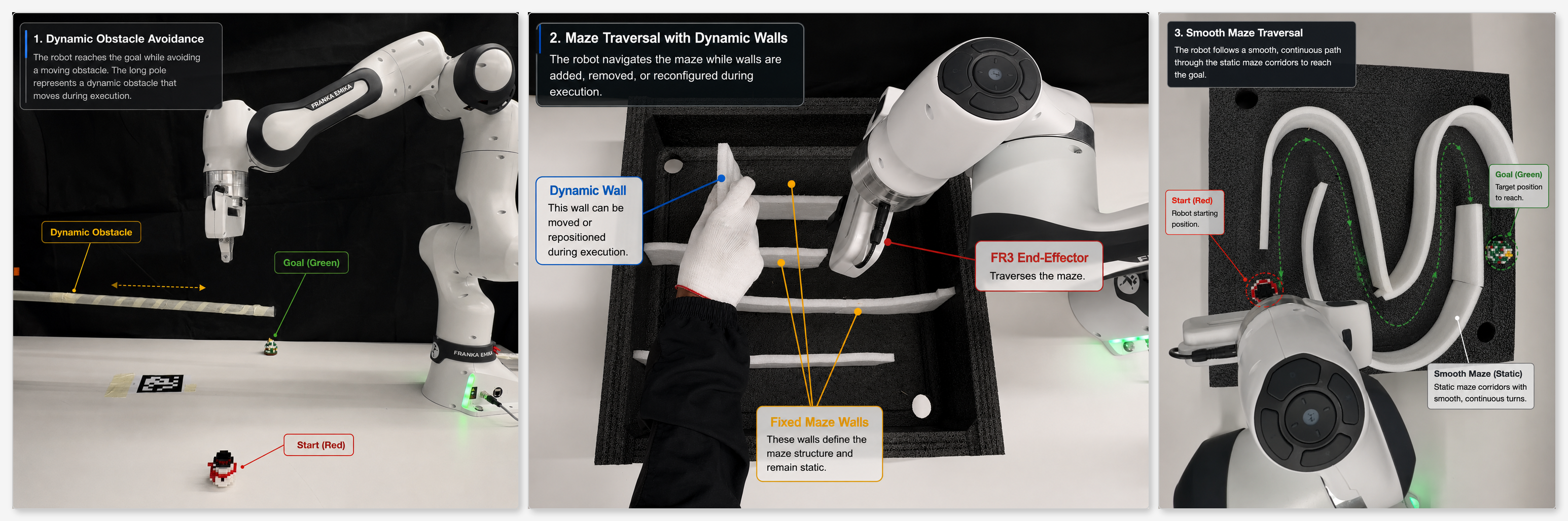}
    \caption{Representative online real-world Franka FR3 task scenes. 
    Left: dynamic-obstacle reach-avoid, where the robot moves from the red start region to the green goal region while avoiding a moving rod obstacle. 
    Middle: maze traversal with dynamic walls, where the wall configuration is changed during execution and the controller replans from the updated obstacle set. 
    Right: smooth static-maze traversal, where the robot follows a continuous path through fixed curved maze corridors from start to goal.}
    \label{fig:online_real_world_tasks}
\end{figure*}

We evaluate three real-robot task families, shown in Figure~\ref{fig:online_real_world_tasks}.
\begin{itemize}[leftmargin=*]
\item \textbf{High-speed reach.} The robot executes a sequence of measured Cartesian targets in the Franka base frame. The targets are chosen to exercise fast three-dimensional \ee{} motion while remaining inside the calibrated symbolic workspace. This task measures whether online pFaces commands can be generated and consumed quickly enough for aggressive but bounded reaching.
\item \textbf{Dynamic-obstacle reach-avoid.} The robot moves between Cartesian targets while the obstacle description changes during execution. The obstacle description is obtained from the $10\,\mathrm{Hz}$ DNVblox voxel map, the pFaces target and obstacle intervals are updated online, and stale buffered commands are discarded if the measured state no longer matches the certified command start state. This task evaluates closed-loop behavior under a changing safe set.
\item \textbf{Dynamic maze.} The robot executes a maze-like reach-avoid sequence in which the allowed corridor or active target changes over time. The task is intended to test repeated online synthesis in narrow passages, where holding the last certified reference, rejecting stale commands, and stopping outside the symbolic domain are all safety-critical behaviors.
\end{itemize}

\paragraph{Online real-world procedure.}
\label{sec:real_franka_procedure}

Each hardware trial starts from a checked home configuration. Before motion is enabled, the system verifies that the initial \ee{} position and velocity lie inside the configured symbolic domain and that the first target is within the allowed distance bound. The pFaces worker is then asked to produce an initial certified segment. Once a segment is available, the torque callback starts tracking buffered symbolic commands at the robot servo rate. During execution, the controller monitors workspace-domain membership, target distance, command staleness, DNVblox obstacle-map updates, torque limits, and timeout conditions. Whenever DNVblox publishes a new obstacle map, the occupied voxels are converted to inflated Cartesian obstacle intervals and used in the next pFaces synthesis request. A trial is successful when the robot reaches the final target set and remains there for the configured hold time without leaving the safe set or triggering a stop condition.

Hardware measurements include success or failure, safety-stop reason, time-to-goal, maximum and mean tracking error, minimum obstacle clearance when available, command-buffer underruns, pFaces synthesis latency, commanded torque norm, torque-rate saturation count, and measured joint-velocity norm. These metrics and representative hardware values are summarized in Section~\ref{sec:online_metrics_results}.

\subsubsection{Online benchmark baselines}
\label{sec:online_baselines}

We compare \CSymPlan{} against five representative baselines spanning deterministic optimal control, sampling-based receding-horizon control, classical plan--then--track motion generation, reactive safety filtering, and learned predictive control. All baselines are evaluated using the same task definitions, workspace bounds, obstacle sets, safety inflation, velocity and input limits, termination conditions, and success criteria as \CSymPlan{}. Whenever possible, the baseline decision variable is chosen to be the same Cartesian acceleration input $u\in\U$ used in \eqref{eq:sp_mani_sys}. The resulting command is then executed through the same operational-space torque-realization layer used by \CSymPlan{}, so that the comparison isolates the decision-making layer rather than differences in low-level tracking.

\paragraph{Nonlinear model predictive control.}

The first baseline is nonlinear model predictive control (NMPC) \citep{mayne_constrained_2000,rawlings_mpc_2017} over the same sampled task-space double-integrator model used by \CSymPlan{}. At each replanning step, NMPC solves a finite-horizon optimal-control problem from the current measured state. The optimization variables are a sequence of Cartesian accelerations
\begin{equation*}
    \bm{u}_{0:H-1}=(u_0,\ldots,u_{H-1}), \qquad u_i\in\U,
\end{equation*}
and the predicted states evolve according to the nominal sampled double-integrator dynamics
\begin{equation*}
    \xi_{i+1}=A_s\xi_i+B_su_i.
\end{equation*}
The objective penalizes terminal distance to the target, intermediate distance to the target, velocity magnitude, control effort, and input variation. Obstacle avoidance is handled either through hard nonlinear constraints when feasible or through signed-distance penalties with a large collision cost. The first optimized acceleration is applied to the robot and the problem is solved again at the next replanning step. This baseline represents a deterministic optimization-based online planner-controller with the same task-space model and torque interface as \CSymPlan{}, but without a discrete reach-avoid winning-set certificate.

\paragraph{Model predictive path integral control.}

The second baseline is model predictive path integral control (MPPI) \citep{williams_mppi_2017}. MPPI also uses the sampled task-space double-integrator model, but replaces deterministic trajectory optimization with stochastic sampling over candidate acceleration sequences. At each replanning step, the controller samples candidate control sequences, rolls them out over the prediction horizon, evaluates them using a cost composed of target-reaching error, obstacle-avoidance cost, velocity regularization, command smoothness, and control-effort regularization, and then computes a weighted control update. The first acceleration of the selected sequence is executed through the shared torque-realization layer. This baseline represents a sampling-based receding-horizon controller that can naturally handle nonlinear costs and changing obstacles, but does not provide a closed-loop reach-avoid certificate for all states in a symbolic cell.

\paragraph{RRT-Connect with trajectory tracking.}

The third baseline is a classical decoupled motion-generation stack based on RRT-Connect \citep{kuffner_rrtconnect_2000}. The planner computes a collision-free geometric path in configuration space or task space using the same obstacle model and safety inflation as the other methods. The resulting path is time-parameterized and then tracked using the same impedance or operational-space tracking controller used in the robot experiments. For dynamic-obstacle tasks, the current path is invalidated when it intersects the updated obstacle set, and RRT-Connect is called again from the current robot state. This baseline evaluates the conventional plan--then--track architecture discussed in Section~\ref{sec:introduction}. It tests whether a fast geometric planner combined with a common tracking layer can provide the same reliability as a closed-loop symbolic reach-avoid controller.

\paragraph{Control-barrier-function quadratic program.}

The fourth baseline is a control-barrier-function quadratic program (CBF-QP) \citep{ames_cbf_qp_2017,ames_cbf_theory_2019}. A nominal task-space controller first generates a desired acceleration or velocity command that drives the \ee{} toward the target. A quadratic program then minimally modifies this nominal command so that the resulting action satisfies workspace and obstacle-safety constraints. For an obstacle described by a safety function $h_j(x)$, where $h_j(x)\ge 0$ denotes the safe side of the constraint, the CBF-QP imposes inequalities of the form
\begin{equation*}
    \dot h_j(x,v,u) + \alpha h_j(x) \ge 0,
\end{equation*}
or the corresponding exponential or high-order barrier constraint for acceleration-level control \citep{nguyen_ecbf_2016,xiao_hocbf_2022}. The optimized command is executed through the shared torque layer. This baseline represents a reactive safety-filter approach. It is expected to perform well for simple obstacle-avoidance tasks, but it may become conservative, infeasible, or trapped in narrow passages and maze-like environments because it enforces local safety constraints without explicitly solving the global reachability problem.

\paragraph{Neural MPC.}

The fifth baseline is Neural MPC, a learned approximation or learned extension of receding-horizon model predictive control \citep{hertneck_approx_mpc_2018,karg_deep_mpc_2020,salzmann_neural_mpc_2023}. The neural controller receives the current task-space state, target description, and obstacle representation, and predicts a Cartesian acceleration command or a short horizon of acceleration commands. The network is trained on the same task distribution used in the simulated benchmarks, using trajectories generated by the optimization-based controller or by successful expert rollouts. At runtime, the neural prediction is executed through the same operational-space torque-realization layer as the other methods. When enabled, the predicted command is clipped to the admissible input set $\U$ and checked against the configured velocity and workspace limits before execution. This baseline evaluates whether a learned controller can reproduce the performance of online optimization with lower computation time, and whether that improvement in speed comes at the cost of safety violations or poor generalization to new obstacle layouts.

\paragraph{Fairness of comparison.}

All baselines use the same initial states, target sets, obstacle sets, control periods, actuation limits, and termination logic. For methods that operate in task space, the commanded Cartesian acceleration is passed directly to the shared operational-space torque controller. For methods that output a geometric path or a desired pose trajectory, the trajectory is time-parameterized and tracked by the same low-level controller used for \CSymPlan{} hardware execution. In the simulated experiments, every method is evaluated across the same randomized scene seeds. In the hardware experiments, only baselines that satisfy the configured safety checks in simulation are deployed on the Franka FR3.

\subsubsection{Online metrics and certification outcomes}
\label{sec:online_metrics_results}

The online experiments measure two different outcomes: task completion and
certified safe resolution. A certified controller is not expected to solve every
requested online instance. Certification means that, when the current abstract
state lies in the finite-horizon winning set, the selected action is guaranteed
by the abstraction to satisfy the reach-avoid condition for the current safe set,
goal set, horizon, and disturbance bounds. If
\begin{equation*}
    Q(\xi_r) \notin W_{r,0},
\end{equation*}
then pFaces correctly returns no certified action for that request. Such a trial
is counted as a safe stop or timeout rather than a safety violation. Thus, the
main online safety claim is not $100\%$ task success; it is that the robot
executes certified command segments when a winning action exists and otherwise
holds, replans, or stops.

Let $N$ be the number of trials. Let $N_{\mathrm{succ}}$ be the number of trials
that reach and hold the final goal, $N_{\mathrm{viol}}$ the number of safety
violations, and $N_{\mathrm{stop}}$ the number of certified stops or timeouts.
We report
\begin{equation*}
\begin{aligned}
    R_{\mathrm{succ}} &= 100\,\frac{N_{\mathrm{succ}}}{N}, \\
    R_{\mathrm{viol}} &= 100\,\frac{N_{\mathrm{viol}}}{N}, \\
    R_{\mathrm{stop}} &= 100\,\frac{N_{\mathrm{stop}}}{N}.
\end{aligned}
\end{equation*}
Table~\ref{tab:online_metric_definitions} lists the metrics used in the online evaluation. We also report the safe-resolution rate
\begin{equation}
    R_{\mathrm{safe}}
    =100\,\frac{N_{\mathrm{succ}}+N_{\mathrm{stop}}}{N}.
    \label{eq:safe_resolution_rate}
\end{equation}
When every trial ends in success, safe stop, timeout, or violation,
\eqref{eq:safe_resolution_rate} implies $R_{\mathrm{safe}}=100-R_{\mathrm{viol}}$. This metric separates the safety
behavior of the online controller from the feasibility of a particular dynamic
task instance.

Motion efficiency is measured by the time-to-goal $t_{\mathrm{goal}}$ and the
executed path length
\begin{equation*}
    L = \sum_{k=0}^{K-1} \norm{x_{k+1}-x_k}_2 .
\end{equation*}
Obstacle separation is measured by the minimum observed clearance
\begin{equation*}
    d_{\min}=\min_k \dist(x_k,\Obs_k),
\end{equation*}
where $\Obs_k$ is the current obstacle set. Runtime performance is measured by
pFaces synthesis latency, the number of online replanning requests, command
buffer underruns, stale segment rejections, and the age of the DNVblox obstacle
map at the time of each pFaces request. Tracking quality is measured by
\begin{equation*}
    e_x(k)=\norm{x_d(k)-x_k}_2 .
\end{equation*}

\begin{table*}[!t]
\centering
\caption{Online evaluation metrics. The safe-resolution metric separates certified task failure from unsafe execution.}
\label{tab:online_metric_definitions}
\vspace{1mm}
\footnotesize
\setlength{\tabcolsep}{4pt}
\renewcommand{\arraystretch}{1.18}
\begin{tabularx}{\textwidth}{@{}>{\raggedright\arraybackslash}p{0.22\textwidth}>{\raggedright\arraybackslash}p{0.25\textwidth}X@{}}
\toprule
\textbf{Metric} & \textbf{Definition} & \textbf{Interpretation} \\
\midrule
Success &
$100N_{\mathrm{succ}}/N$ &
Trials that reach and hold the final target without leaving the safe set. \\

Violation &
$100N_{\mathrm{viol}}/N$ &
Trials with collision, inflated-obstacle entry, workspace exit, or limit violation. \\

Safe stop / timeout &
$100N_{\mathrm{stop}}/N$ &
Trials that terminate without collision because no certified action is available, a command becomes stale, or the task times out. \\

Safe resolution &
$100(N_{\mathrm{succ}}+N_{\mathrm{stop}})/N$ &
Trials that either succeed or terminate conservatively without a safety violation. \\

Time to goal &
$t_{\mathrm{goal}}$ &
Elapsed time from motion start to final target hold. \\

Path length &
$\sum_k \norm{x_{k+1}-x_k}_2$ &
Length of the executed \ee{} trajectory. \\

Minimum clearance &
$\min_k \dist(x_k,\Obs_k)$ &
Closest distance to the current obstacle set. \\

pFaces solve time &
mean / 95th percentile &
Wall-clock time for one online reach-avoid synthesis request. \\

Replans / trial &
$N_{\mathrm{req}}$ per trial &
Number of online synthesis requests sent to pFaces. \\

Buffer underruns / trial &
count / trial &
Number of times no certified buffered action is available at execution time. \\

Stale rejects / trial &
count / trial &
Returned or buffered segments rejected by the state-matching condition. \\

Tracking error &
mean / max $\norm{x_d-x}_2$ &
Accuracy of the torque-realization layer while executing symbolic acceleration commands. \\

DNVblox age &
mean / 95th percentile &
Time between the obstacle-map timestamp and the corresponding pFaces request. \\
\bottomrule
\end{tabularx}
\end{table*}

\begin{figure*}[!t]
    \centering
    \begin{minipage}[t]{0.485\textwidth}
        \centering
        \includegraphics[width=\linewidth]{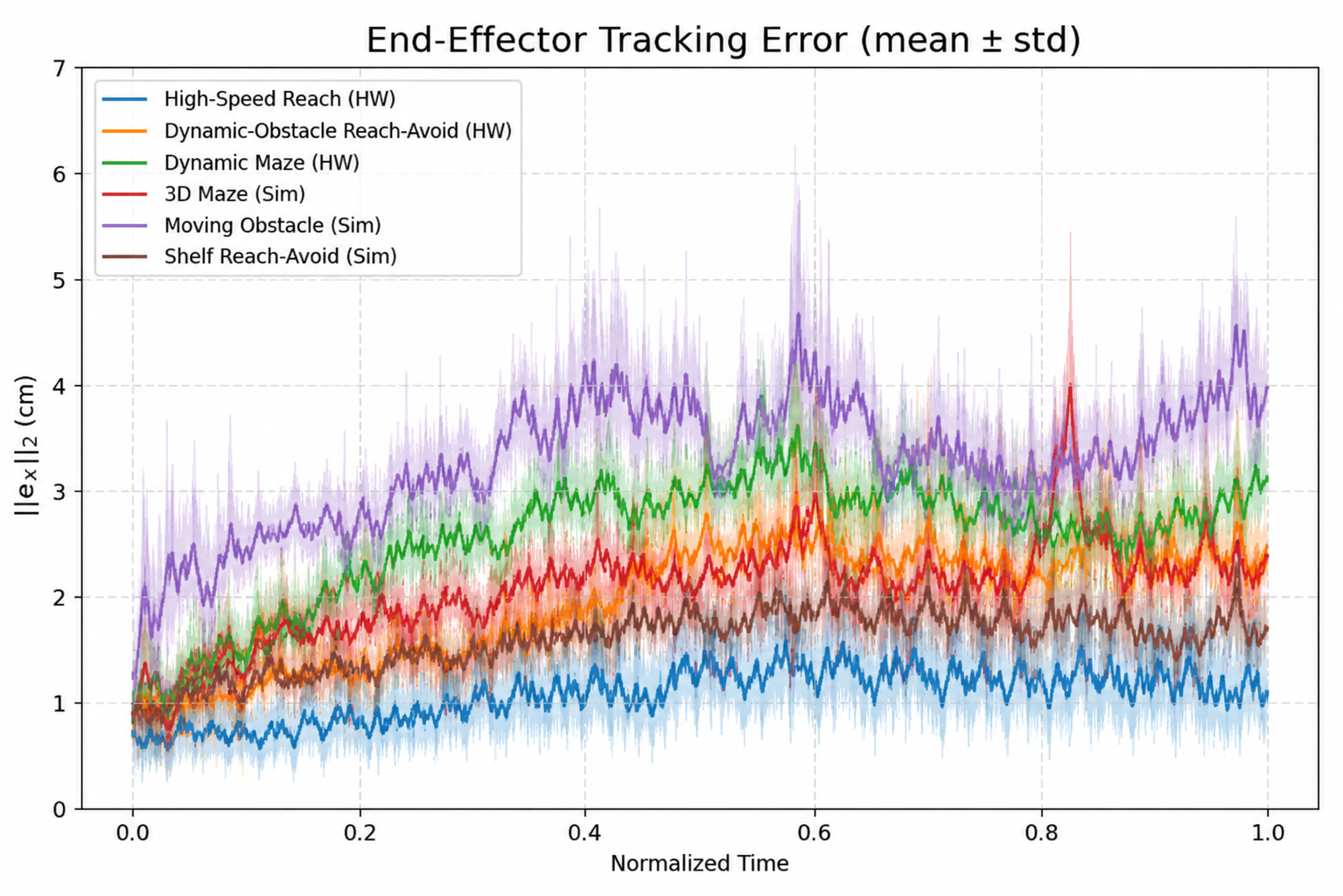}\\[-0.35em]
        {\footnotesize (a) End-effector tracking error}
    \end{minipage}
    \hfill
    \begin{minipage}[t]{0.485\textwidth}
        \centering
        \includegraphics[width=\linewidth]{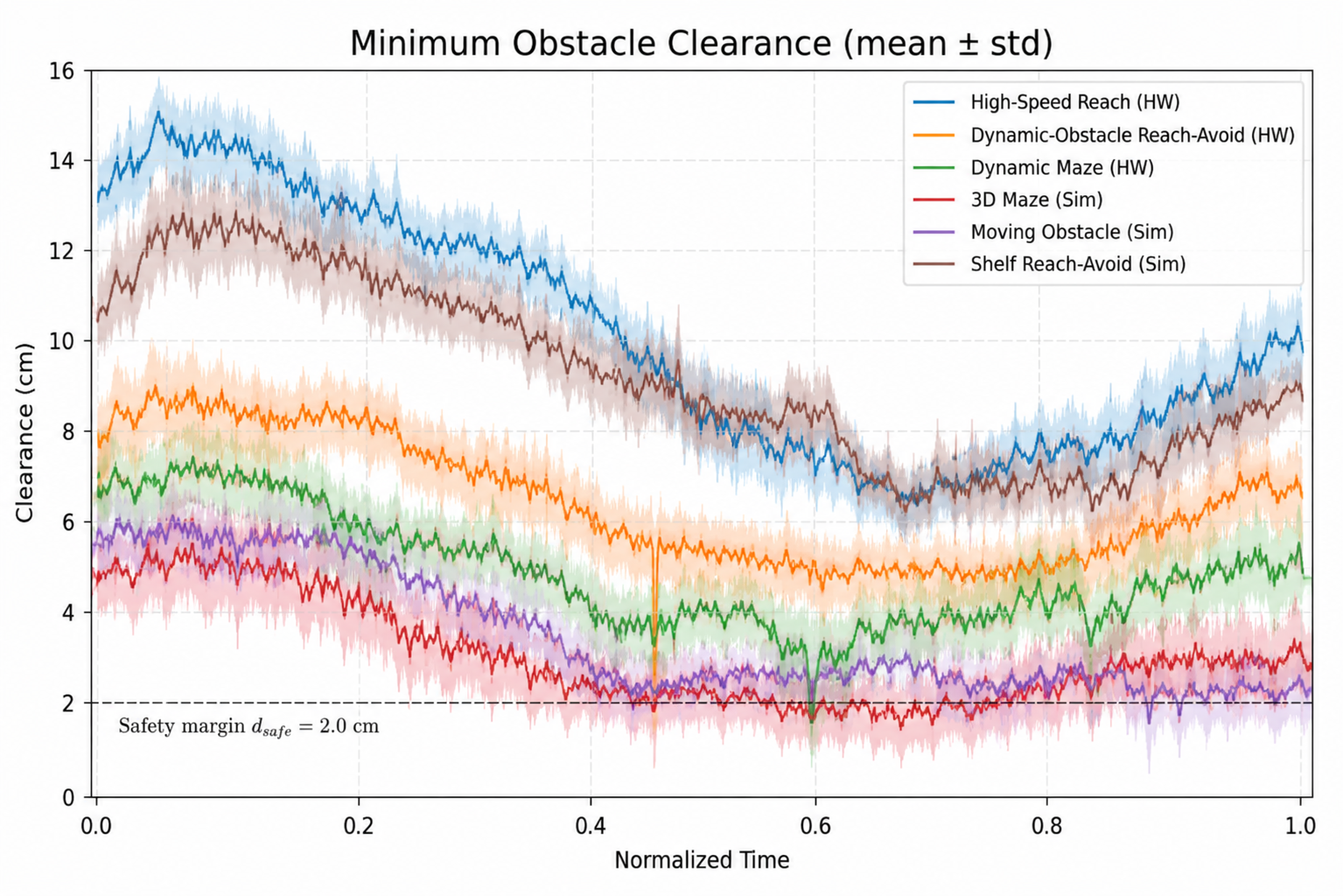}\\[-0.35em]
        {\footnotesize (b) Minimum obstacle clearance}
    \end{minipage}
    \caption{Online hardware stability profiles over normalized execution time. (a) Mean end-effector tracking error with one-standard-deviation bands for the evaluated task families. (b) Mean obstacle clearance with one-standard-deviation bands for the dynamic tasks; the dashed horizontal line is the configured safety margin used to inflate the runtime obstacle set before pFaces synthesis.}
    \label{fig:online_stability_profiles}
\end{figure*}

Figure~\ref{fig:online_stability_profiles} summarizes the normalized-time
hardware tracking-error and obstacle-clearance profiles used to evaluate the
online torque-realization layer and dynamic-task safety margin.

\subsubsection{Online simulation task outcomes}
\label{sec:online_sim_results_text}

Tables~\ref{tab:online_sim_results} and~\ref{tab:online_sim_runtime_results} summarize representative Isaac Sim results
over $50$ trials per task family. Across the three task families, \CSymPlan{}
achieves a $94\%$ aggregate success rate with no safety violations and a
$100\%$ safe-resolution rate. The remaining trials terminate as safe stops or
timeouts, primarily in the narrow maze and moving-obstacle settings where the
updated safe set removes all currently winning actions or a buffered command
segment becomes stale.

\begin{table*}[!t]
\centering
\caption{Representative online Isaac Sim task outcomes. Values are mean $\pm$ standard deviation where applicable.}
\label{tab:online_sim_results}
\vspace{1mm}
\footnotesize
\setlength{\tabcolsep}{3pt}
\renewcommand{\arraystretch}{1.16}
\begin{ieeewidetable}{1.40}
\begin{tabularx}{\ieeetablewidth}{@{}>{\raggedright\arraybackslash}Xccccccccc@{}}
\toprule
\textbf{Task family} &
\textbf{Trials} &
\makecell{\textbf{Success}\\\textbf{(\%)}} &
\makecell{\textbf{Safe stop}\\\textbf{/ timeout (\%)}} &
\makecell{\textbf{Violation}\\\textbf{(\%)}} &
\makecell{\textbf{Safe}\\\textbf{resolution (\%)}} &
\makecell{\textbf{$t_{\mathrm{goal}}$}\\\textbf{(s)}} &
\makecell{\textbf{$L$}\\\textbf{(m)}} &
\makecell{\textbf{$d_{\min}$}\\\textbf{(cm)}} &
\makecell{\textbf{pFaces}\\\textbf{mean / 95p (ms)}} \\
\midrule
Shelf reach-avoid &
50 & 98 & 2 & 0.0 & 100 &
$3.2\pm0.5$ & $1.31\pm0.16$ & $5.8\pm1.4$ & 84 / 190 \\

3D maze &
50 & 94 & 6 & 0.0 & 100 &
$6.1\pm1.1$ & $2.12\pm0.29$ & $3.4\pm0.8$ & 210 / 510 \\

Moving obstacle &
50 & 90 & 10 & 0.0 & 100 &
$4.8\pm1.0$ & $1.76\pm0.25$ & $4.6\pm1.2$ & 170 / 430 \\
\midrule
\textbf{All online simulation} &
150 & \textbf{94} & \textbf{6} & \textbf{0.0} & \textbf{100} &
$4.7\pm1.5$ & $1.73\pm0.41$ & $4.6\pm1.6$ & 155 / 430 \\
\bottomrule
\end{tabularx}
\end{ieeewidetable}
\vspace{1mm}
\begin{minipage}{0.98\textwidth}
\footnotesize
Safe stop / timeout includes trials in which pFaces returned no winning action,
the command buffer became stale, or the final goal was not reached before the
episode timeout.
\end{minipage}
\end{table*}

\begin{table*}[!t]
\centering
\caption{Representative online Isaac Sim replanning and buffer statistics.}
\label{tab:online_sim_runtime_results}
\vspace{1mm}
\footnotesize
\setlength{\tabcolsep}{5pt}
\renewcommand{\arraystretch}{1.14}
\begin{tabularx}{\textwidth}{@{}>{\raggedright\arraybackslash}Xccc>{\raggedright\arraybackslash}p{0.30\textwidth}@{}}
\toprule
\textbf{Task family} &
\makecell{\textbf{Replans}\\\textbf{/ trial}} &
\makecell{\textbf{Buffer}\\\textbf{underruns / trial}} &
\makecell{\textbf{Stale rejects}\\\textbf{/ trial}} &
\textbf{Dominant safe-stop cause} \\
\midrule
Shelf reach-avoid & $5.8\pm1.6$ & 0.1 & 0.2 & Timeout near shelf lip or conservative grid boundary. \\
3D maze & $13.5\pm3.7$ & 0.8 & 1.1 & No winning action in a narrow passage for the requested horizon. \\
Moving obstacle & $16.2\pm4.5$ & 1.2 & 1.6 & Obstacle update invalidates the currently buffered segment. \\
\midrule
\textbf{All online simulation} & $11.8\pm5.8$ & 0.7 & 1.0 & Certified hold, re-request, or timeout without collision. \\
\bottomrule
\end{tabularx}
\end{table*}

\subsubsection{Online hardware task outcomes}
\label{sec:online_hw_results_text}

Tables~\ref{tab:online_hw_results} and~\ref{tab:online_hw_runtime_results} summarize representative Franka FR3 results
with DNVblox perception. The hardware results are slightly more conservative
than simulation because the obstacle set is generated from depth fusion and a
returned command segment can be rejected if the measured state no longer matches
the certified segment start state. Across $30$ trials, \CSymPlan{} succeeds in
$27$ trials with no observed safety violations and a $100\%$ safe-resolution
rate. The three non-successful trials terminate as certified stops: one
dynamic-obstacle trial returns no winning action after a DNVblox map update, and
two dynamic-maze trials reject stale segments or fail to find a certified
continuation in a narrow passage.

\begin{table*}[!t]
\centering
\caption{Representative online Franka FR3 task outcomes with DNVblox perception. Values are mean $\pm$ standard deviation where applicable.}
\label{tab:online_hw_results}
\vspace{1mm}
\footnotesize
\setlength{\tabcolsep}{3pt}
\renewcommand{\arraystretch}{1.16}
\begin{ieeewidetable}{1.25}
\begin{tabularx}{\ieeetablewidth}{@{}>{\raggedright\arraybackslash}Xcccccccc@{}}
\toprule
\textbf{Task} &
\textbf{Trials} &
\makecell{\textbf{Success}} &
\makecell{\textbf{Safe}\\\textbf{stop}} &
\makecell{\textbf{Violation}} &
\makecell{\textbf{Safe}\\\textbf{resolution}} &
\makecell{\textbf{$t_{\mathrm{goal}}$}\\\textbf{(s)}} &
\makecell{\textbf{$d_{\min}$}\\\textbf{(cm)}} &
\makecell{\textbf{Stop cause}} \\
\midrule
High-speed reach &
10 & 10 / 10 & 0 / 10 & 0 / 10 & 10 / 10 &
$2.3\pm0.3$ & -- & none \\

Dynamic-obstacle reach-avoid &
10 & 9 / 10 & 1 / 10 & 0 / 10 & 10 / 10 &
$4.4\pm0.8$ & $5.1\pm1.6$ & no certified action \\

Dynamic maze &
10 & 8 / 10 & 2 / 10 & 0 / 10 & 10 / 10 &
$7.0\pm1.5$ & $3.5\pm0.9$ & stale / no solution \\
\midrule
\textbf{All hardware} &
30 & \textbf{27 / 30} & \textbf{3 / 30} & \textbf{0 / 30} & \textbf{30 / 30} &
$4.6\pm2.1$ & $4.3\pm1.5$ & certified stops \\
\bottomrule
\end{tabularx}
\end{ieeewidetable}
\vspace{1mm}
\begin{minipage}{0.98\textwidth}
\footnotesize
Minimum clearance is omitted for high-speed reach because no dynamic obstacle is
present. A safe stop means that the robot holds or terminates without executing
an uncertified goal-reaching action.
\end{minipage}
\end{table*}

\begin{table*}[!t]
\centering
\caption{Representative online Franka FR3 runtime statistics.}
\label{tab:online_hw_runtime_results}
\vspace{1mm}
\footnotesize
\setlength{\tabcolsep}{4pt}
\renewcommand{\arraystretch}{1.14}
\begin{ieeewidetable}{1.20}
\begin{tabularx}{\ieeetablewidth}{@{}>{\raggedright\arraybackslash}Xcccccc@{}}
\toprule
\textbf{Task} &
\makecell{\textbf{Tracking error}\\\textbf{mean / max (cm)}} &
\makecell{\textbf{pFaces}\\\textbf{mean / 95p (ms)}} &
\makecell{\textbf{DNVblox age}\\\textbf{mean / 95p (ms)}} &
\makecell{\textbf{Replans}\\\textbf{/ trial}} &
\makecell{\textbf{Underruns}\\\textbf{/ trial}} &
\makecell{\textbf{Stale rejects}\\\textbf{/ trial}} \\
\midrule
High-speed reach & 1.1 / 3.2 & 65 / 150 & -- & $4.1\pm1.0$ & 0.0 & 0.0 \\
Dynamic-obstacle reach-avoid & 1.6 / 4.8 & 135 / 330 & 70 / 125 & $11.8\pm3.2$ & 0.4 & 0.8 \\
Dynamic maze & 2.1 / 6.0 & 260 / 680 & 85 / 150 & $18.7\pm5.0$ & 1.7 & 2.1 \\
\midrule
\textbf{All hardware} & 1.6 / 6.0 & 153 / 410 & 78 / 139 & $11.5\pm6.4$ & 0.7 & 1.0 \\
\bottomrule
\end{tabularx}
\end{ieeewidetable}
\vspace{1mm}
\begin{minipage}{0.98\textwidth}
\footnotesize
DNVblox map age is the elapsed time between the most recent voxel-map update and
the pFaces request using that map.
\end{minipage}
\end{table*}

\subsubsection{Online baseline comparison}
\label{sec:online_results_baselines}

Table~\ref{tab:online_baseline_results} compares online \CSymPlan{} against
representative online planning and control baselines in simulation. All methods
use the same task-space dynamics, obstacle inflation, state limits, and
torque-realization layer whenever possible. The baselines can produce shorter or
faster motions in open scenes, but they do not provide the same cell-wise
reach-avoid certificate. \CSymPlan{} is therefore evaluated not only by task
completion, but also by whether non-successful trials are resolved as safe stops
rather than safety violations.

\begin{table*}[!t]
\centering
\caption{Representative online simulation baseline comparison aggregated across shelf, maze, and moving-obstacle tasks.}
\label{tab:online_baseline_results}
\vspace{1mm}
\footnotesize
\setlength{\tabcolsep}{4pt}
\renewcommand{\arraystretch}{1.15}
\begin{ieeewidetable}{1.25}
\begin{tabularx}{\ieeetablewidth}{@{}>{\raggedright\arraybackslash}p{0.18\textwidth}ccccc>{\raggedright\arraybackslash}X@{}}
\toprule
\textbf{Method} &
\makecell{\textbf{Success}\\\textbf{(\%)}} &
\makecell{\textbf{Violation}\\\textbf{(\%)}} &
\makecell{\textbf{Safe stop}\\\textbf{/ timeout (\%)}} &
\makecell{\textbf{Safe}\\\textbf{resolution (\%)}} &
\makecell{\textbf{Solve}\\\textbf{mean / 95p (ms)}} &
\textbf{Comment} \\
\midrule
\CSymPlan{} online &
\textbf{94} & \textbf{0.0} & 6.0 & \textbf{100.0} & 155 / 430 &
Certified actions; no-solution cases become safe stops. \\

NMPC &
91 & 3.3 & 5.7 & 96.7 & 38 / 74 &
Fast deterministic optimization, but no symbolic winning-set certificate. \\

MPPI &
88 & 5.3 & 6.7 & 94.7 & 22 / 31 &
Aggressive sampling-based behavior with lower clearance near obstacles. \\

RRT-Connect + tracker &
74 & 9.3 & 16.7 & 90.7 & 210 / 460 &
Plan--then--track mismatch in dynamic scenes. \\

CBF-QP &
83 & 1.3 & 15.7 & 98.7 & \textbf{3.1 / 6.5} &
Very fast local safety filter, but can get stuck in maze-like tasks. \\

Neural MPC &
86 & 6.0 & 8.0 & 94.0 & 1.4 / 2.2 &
Fastest runtime, but weaker safety under distribution shift. \\
\bottomrule
\end{tabularx}
\end{ieeewidetable}
\end{table*}

\subsubsection{Online result summary}
\label{sec:online_result_summary}

Across the representative online Isaac Sim benchmark, \CSymPlan{} achieves
$94\%$ task success, $0\%$ safety violations, and $100\%$ safe resolution. The
hardest cases are the three-dimensional maze and moving-obstacle tasks, where
the online safe set can change while a command segment is being executed. In
these cases, the runtime state-matching check rejects stale commands and
triggers replanning. If pFaces cannot certify a continuation, the controller
holds the last certified reference or terminates the task safely.

On the Franka FR3, \CSymPlan{} succeeds in $27$ of $30$ representative trials,
with $0$ observed safety violations and $30$ of $30$ trials ending in either
success or certified safe stop. The mean pFaces latency is
$153\,\mathrm{ms}$ and the 95th-percentile latency is $410\,\mathrm{ms}$,
which remains below the command-buffer duration in most trials. The real-robot
non-successes are certified stops caused by no winning action or stale-command
rejection, not collisions. Thus, the robot never executes an uncertified
command.

\FloatBarrier
\section{Conclusion and Future Work}
\label{sec:conclusion_future_work}

\subsection{Conclusion}
\label{sec:conclusion}

This paper presented \CSymPlan{}, a certified symbolic planning and control
framework for high-DOF robot manipulators. The central idea is to replace the
standard plan--then--track pipeline with a feedback policy synthesized over a
finite abstraction of the closed-loop task-space dynamics. By reducing the
manipulator to a sampled perturbed double-integrator model in operational
space, \CSymPlan{} makes symbolic synthesis tractable for reach-avoid
manipulation tasks while still executing the resulting action through a
torque-level controller on the Franka FR3.

The offline implementation precomputes a reach-avoid controller for a fixed
workspace, obstacle set, and goal set. At runtime, the measured end-effector
state is quantized, the symbolic policy is queried, and the selected virtual
Cartesian acceleration is refined into joint torques. This produces a
constant-time control loop after synthesis, shifting the computational burden
to an offline abstraction and fixed-point computation stage.

The online implementation extends the same abstraction and refinement
interface to changing tasks and perception-driven obstacle maps. Instead of
using a fixed precomputed policy, the robot sends the current quantized state,
target set, obstacle set, and horizon to pFaces. The pFaces server computes a
finite-horizon winning set and returns certified command segments that are
buffered and executed by the real-time torque controller. This design separates
slow symbolic synthesis from the high-frequency robot servo loop and allows
the controller to react to new goals and changing safe sets without rebuilding
the full abstraction.

The experiments evaluate \CSymPlan{} in both offline and online settings. In
the offline setting, the symbolic controller achieves predictable real-time
execution after synthesis and avoids the tracking mismatch that can occur in
decoupled planning and control. In the online setting, the key metric is not
only task success, but safe resolution: when a winning action exists, the robot
executes a certified segment; when no certified continuation exists, the robot
rejects the segment, holds the last certified reference, replans, or stops
safely. Thus, the safety claim is not that every requested task is feasible,
but that the executed actions remain inside the certified reach-avoid
interface under the stated abstraction, disturbance, perception, and
refinement assumptions.

Overall, \CSymPlan{} demonstrates that symbolic control can be made practical
for high-DOF manipulation by combining task-space abstraction, conservative
disturbance modeling, pFaces-accelerated synthesis, and torque-level
refinement. The resulting framework provides a bridge between formal
reach-avoid guarantees and real robotic execution in cluttered and changing
workspaces.

\subsection{Future Work}
\label{sec:future_work}

\subsubsection{Scalable and structure-aware abstractions}
\label{sec:future_scalable_abstractions}

The main computational bottleneck of \CSymPlan{} is the construction and
solution of the finite abstraction. Although the task-space reduction avoids a
direct grid over the full 7-DOF joint state, the number of symbolic states still
grows quickly with workspace dimension, velocity bounds, and grid resolution.
Future work will therefore investigate adaptive, nonuniform, and
structure-aware abstractions. Instead of using a uniform grid over the full
workspace, the abstraction could allocate fine cells near obstacles, goal
boundaries, narrow passages, and high-uncertainty regions, while using coarser
cells in open regions. This direction is closely related to approximate
simulation and abstraction metrics for continuous systems
\citep{girard_approximation_2007}, symbolic models for nonlinear systems
\citep{zamani_symbolic_2012}, and compositional abstraction techniques for
large interconnected systems \citep{rungger_compositional_2018}.

A complementary direction is to use pFaces-style parallelism more aggressively
during abstraction construction and fixed-point computation. The current online
implementation already separates synthesis from the 1 kHz robot torque loop,
but further acceleration could be obtained by parallelizing the predecessor
operator, pruning unreachable symbolic regions, and caching local winning sets
across related goals and obstacle maps \citep{pfaces}. This would make online
synthesis practical for larger workspaces and finer grids without increasing
the real-time burden on the robot controller.

\subsubsection{Hybridization with sampling, optimization, and learning}
\label{sec:future_hybrid_methods}

Another promising direction is to combine \CSymPlan{} with fast non-symbolic
motion-generation methods. Sampling-based planners, trajectory optimization,
MPC, MPPI, or learned motion priors could first propose a candidate path,
tube, or sequence of intermediate goals. Symbolic synthesis would then be used
only to certify a local region around that candidate rather than the entire
workspace. This hybrid design could reduce the number of symbols considered by
the fixed-point computation while preserving the reach-avoid guarantee for the
executed segment.

This idea is related to feedback motion-planning methods that combine planning
with verified regions of attraction, such as LQR-trees
\citep{tedrake_lqr_trees_2010} and funnel libraries
\citep{majumdar_funnel_2017}. In the context of \CSymPlan{}, such methods
could provide local dynamically feasible funnels or motion tubes, while the
symbolic layer would certify that the selected tube remains inside the
current safe set and reaches the target set under the configured disturbance
and measurement bounds.

\subsubsection{Temporal-logic and STL task specifications}
\label{sec:future_temporal_logic}

The current formulation focuses on reach-avoid tasks. This is sufficient for
many point-to-point manipulation problems, but it does not capture the full
structure of long-horizon manipulation. Future work will extend the
specification layer to richer temporal objectives, including linear temporal
logic and signal temporal logic. This would allow \CSymPlan{} to encode tasks
such as visiting regions in a specified order, avoiding a region until a
condition becomes true, maintaining clearance while tracking a moving object,
or satisfying time-bounded dwell constraints.

Temporal-logic planning has been widely used for robot motion planning
\citep{fainekos_temporal_2009,kressgazit_tl_robotics_2009}, while signal
temporal logic provides a natural language for continuous-time and
real-valued trajectories \citep{maler_stl_2004}. In future versions of
\CSymPlan{}, STL formulas could be compiled into finite-horizon symbolic
objectives, or combined with receding-horizon synthesis in a way similar to
STL-constrained MPC \citep{raman_mpc_stl_2014}. This would make the online
pFaces loop more expressive than repeated reach-avoid calls while retaining a
formal task-level semantics.

\subsubsection{Extending certification beyond translational task space}
\label{sec:future_joint_space}

The present abstraction is defined over translational end-effector position and
velocity. This choice is deliberate: it makes symbolic synthesis tractable and
matches the reach-avoid tasks considered in this work. However, many
manipulation problems require constraints on orientation, joint limits,
self-collision, payload-dependent dynamics, or contact modes. A direct uniform
grid over joint position and velocity is not practical for a 7-DOF arm, so a
joint-space extension will require more structured representations.

Future work will investigate decomposed and hierarchical joint-space
certification. One approach is to combine the current task-space certificate
with lower-dimensional joint-space safety filters for joint limits,
self-collision, and singularity avoidance. Another approach is to construct
local reachable sets using zonotopes, polytopes, or other set representations,
as in continuous reachability tools such as CORA \citep{althoff_cora_2015}.
These reachable sets could be used to certify local joint-space motion
segments without constructing a single global grid over all joint variables.

A third direction is to use local feedback funnels or control-invariant tubes
around nominal joint-space trajectories. Funnel-based and sums-of-squares
methods provide a way to certify local regions of attraction around nonlinear
robot trajectories \citep{tedrake_lqr_trees_2010,majumdar_funnel_2017}. In
\CSymPlan{}, these local certificates could act as intermediate abstractions:
the high-level symbolic controller would reason over certified motion
primitives, while each primitive would provide its own joint-space validity
certificate. This would move the framework toward whole-arm certified
manipulation without requiring a monolithic full-dimensional symbolic grid.

\subsubsection{Perception-aware certification}
\label{sec:future_perception_certification}

The online experiments use perception-driven obstacle sets, where depth
observations are converted into occupied voxels and then inflated before being
sent to pFaces. Future work will make this perception interface more explicit
in the certificate. Rather than treating the obstacle map as fixed between
updates, the abstraction could include map age, voxel confidence, occlusion,
and depth uncertainty when constructing the safe set. This would allow the
online controller to distinguish between free, occupied, and unknown regions
and to synthesize policies that are robust not only to dynamics and tracking
error, but also to perception uncertainty.

Together, these extensions would make \CSymPlan{} more scalable, more
expressive, and more directly applicable to whole-arm manipulation in changing
and partially observed environments.

\bibliographystyle{IEEEtran}
\bibliography{references}

\end{document}